\documentclass[journal]{IEEEtran}

\usepackage{xcolor,soul,framed} 

\colorlet{shadecolor}{yellow}
\usepackage[pdftex]{graphicx}
\graphicspath{{../pdf/}{../jpeg/}}
\DeclareGraphicsExtensions{.pdf,.jpeg,.png}

\usepackage[cmex10]{amsmath}
\usepackage{array}
\usepackage{mdwmath}
\usepackage{booktabs} 
\usepackage{mdwtab}
\usepackage{eqparbox}
\usepackage{amssymb}
\usepackage{url}
\usepackage{algorithm}
\usepackage{algorithmic}
\usepackage{cite}
\usepackage{comment}
\usepackage{multirow}
\usepackage{amsmath}
\usepackage{tabularx}
\usepackage{adjustbox}
\usepackage{gensymb}
\usepackage[caption=false,font=footnotesize]{subfig}
\begin{document}
\bstctlcite{IEEEexample:BSTcontrol}
    \title{Adaptive Gait Generation for Multi-Terrain Exoskeletons via Constrained Kernelized Movement Primitives}
  \author{Edoardo~Trombin,
    Miroljub~Mihailovic,
    Matheus~Henrique~Ferreira~Moura,
    Luca~Tonin,
    \\Emanuele~Menegatti
    and Stefano~Tortora%
  \thanks{All authors are with the Department of Information Engineering, University of Padua, 35131 Padua, Italy. LT, EM and ST are also with the Padova Neuroscience Center, University of Padua (corresponding author: Stefano Tortora, e-mail: stefano.tortora@unipd.it).}
  }

\maketitle

\begin{abstract}
Lower limb exoskeletons (LLEs) present the potential to make motor-impaired individuals walk again. Their application in real-world environments is still limited by the lack of effective adaptive gait planning. Indeed, current exoskeletons are meant to walk only on a flat and even terrain. Generating environment-aware, physiologically consistent gait trajectories in real-time is an open challenge. To overcome this, we propose a novel Kernelized Movement Primitives (KMP)-based framework for adaptive gait generation (AGG) across multiple indoor terrains. The proposed approach learns a probabilistic representation of human gait in both the joint and task spaces from a limited number of human demonstrations, representing natural gait characteristics and ensuring kinematic feasibility. In addition, the learned trajectories are adapted using environmental information extracted from an onboard RGB-D camera by treating the AGG as a linearly constrained optimization problem with via-points. The proposed method has been thoroughly validated first in simulations for gait generation in different scenarios, such as flat-ground walking, slopes, stairs, and obstacles crossing. Finally, the effectiveness and robustness of the method have been demonstrated with experiments on a commercial LLE in real-world scenarios. The results obtained demonstrate the feasibility of an environment-aware gait planning system for a new generation of intelligent lower limb exoskeletons for assisting people with disabilities in their every-day life. 
\end{abstract}


%
\IEEEpeerreviewmaketitle


\section{Introduction}
\IEEEPARstart{L}{\lowercase{ower}} limb exoskeletons (LLEs) are increasingly investigated as assistive devices for individuals with motor impairments: by providing mechanical support and powered actuation, they allow users to regain the ability to stand and perform walking tasks \cite{pamungkas2019overview, de_la_tejera_systematic_2021}.

Despite several years of research, the use of these devices is limited to clinical environments. Nevertheless, the everyday use of exoskeletons holds great potential to enhance users’ independence and decrease the incidence of secondary health conditions, thereby improving the overall quality of life for individuals with lower-limb disabilities \cite{van2024improvement}. One of the key reasons for this limitation can be attributed to the fact that most of the existing LLEs still relies on pre-defined trajectories that are only partially customizable to user-specific parameters and walking speed \cite{strausser_mobile_2012,esquenazi_rewalk_2012}. While these systems are suitable for controlled environments, they cannot cope with scenarios requiring a run-time adaptation of the walking pattern. This limits applicability in real-world scenarios where users could encounter a wide array of environmental elements: the presence of slopes, stairs, and obstacles along the walking path poses a nontrivial challenge for current exoskeletons. 

To address this challenge, environment-adaptive gait planning represents a recent research trend in LLEs that allows the exoskeleton to generate adaptive gait trajectories based on the specific environmental characteristics (e.g., type of terrain, presence of obstacles) \cite{yao2024advancements}. Overall, these new exoskeleton systems are characterized by two complementary components: (i) a vision module based on onboard exteroceptive sensors (e.g., range sensors, RGB-D cameras) to analyze and identify upcoming terrain features, and (ii) an Adaptive Gait Generation (AGG) module implementing algorithms for computing at runtime the exoskeleton's joint trajectories based on the terrain. While perception capabilities of LLEs are steadily improving \cite{wang_review_2024}, the bottleneck remains gait generation: producing trajectories that are both adaptive to environmental variations while respecting the constraints required for safe exoskeleton use is a complex task whose solution requires both robustness and adaptability.

Currently, the most common AGG methods for LLEs are based on analytical solutions employing geometry-based models solved with simple interpolation \cite{liu_vision-assisted_2021,ren_fast_2020, ren_-line_2018, mohamad2023online}, or through more advanced stochastic optimization \cite{mohamad_minimum-time_2023,kazemi_real-time_2019, mohamad_optimization_2024,trombin2024environment}, and nonlinear optimization \cite{li_human---loop_2024, hua_vision_2022} techniques. Also, Reinforcement Learning (RL) has been explored for generating adaptive gait to avoid the explicit definition of a model \cite{guo_research_2024, trombin_environment-adaptive_2025}. More recently, data-driven solutions based on imitation learning have been proposed to learn the complexity of the locomotion task from human demonstrations. Among these imitative solutions, Dynamic Movement Primitives (DMPs) \cite{kimura_dynamic_2006, ijspeert_dynamical_2013, wang_review_2024} and its variants became particularly popular for generating adaptive gait in LLEs as they provide an elegant formulation for encoding complex motion dynamics, preserving e.g. the physiological characteristics of the gait, while guaranteeing at the same time a high degree of adaptability \cite{huang_adaptive_2020, chen_learning_2019}. In particular, Kernelized Movement Primitives (KMPs) \cite{huang2019kernelized} represent one of the latest variants of the original DMP framework that allows the encoding of the trajectories as probability distributions, and the generation of new trajectories that preserve the statistical properties of the learned trajectories. Thanks to these properties, KMPs have been extensively studied for the control of robotic manipulators \cite{liu2023variable,liu2025collaborative, zhao2025reactive}, robotic hands for dexterous grasping \cite{katyara2021leveraging,wang2024humanoid}, humanoid robots \cite{colome2014dimensionality}, and wearable robots \cite{liu2024human}. 
Despite these latest advancements, the application of KMP to LLE's locomotion is little explored, as detailed in Section \ref{sec:related}, and a full environment-aware AGG system for exoskeletons for multi-terrain walking is still missing.

In this work, we present a novel KMP-based AGG method for multi-terrain adaptive gait planning in lower-limb exoskeletons. In particular, the proposed method learns a probabilistic representation of gait trajectories directly from a few human demonstrations, both in the task space (i.e., swing foot trajectory) and in the joint space (i.e., support leg angular trajectories). By applying linear constraints based on the environmental characteristics measured by an RGB-D camera mounted on the exoskeleton, the proposed method is capable of adapting the learned trajectories at run time to different indoor terrains (level-ground, slopes, and stairs) and to the presence of obstacles hindering the walking path. The efficacy and robustness of the proposed approach were validated both in simulated scenarios and with a real commercial exoskeleton.
The main contributions of this work are summarized as follows:
\begin{itemize}
    \item definition of a novel KMP framework to learn and generate gait trajectories simultaneously in task and joint spaces, preserving natural gait characteristics while ensuring adaptability to the environment;
    \item development of an environment-aware adaptive gait planning framework integrating RGB-D perception into the KMP formulation, tackling the AGG as a linearly constrained optimization problem;
    \item extensive validation of the proposed method, both in simulation and on a commercial lower-limb exoskeleton across multiple terrains (e.g., stairs, slopes, and obstacle crossing), demonstrating robust and adaptive performance in real-world conditions.
\end{itemize}

\section{Related Work}
\label{sec:related}
\begin{figure*}[t]
\centerline{\includegraphics[width=0.85\linewidth]{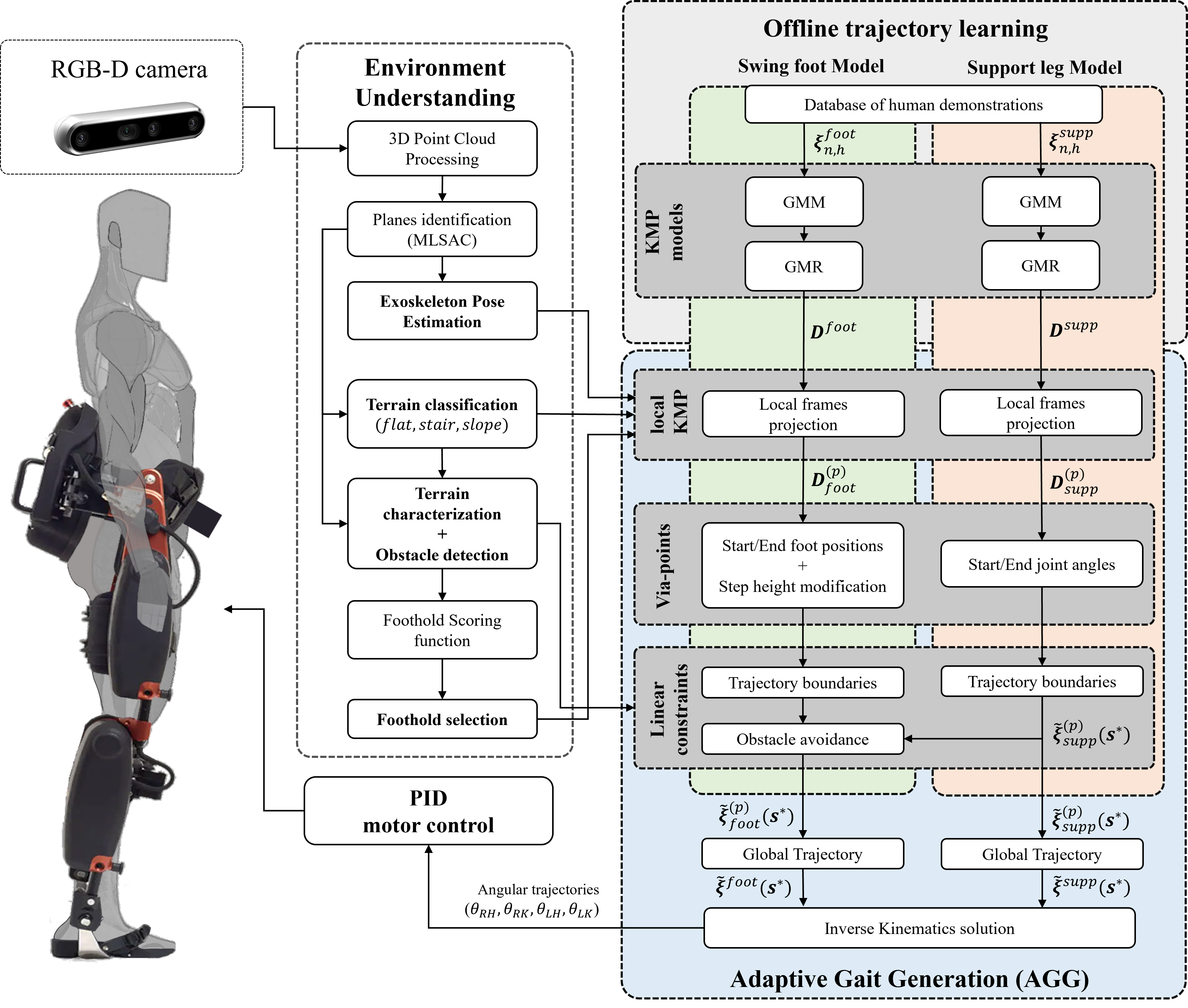}}
\caption{Overview of the proposed solution for environment-adaptive gait generation. The \textit{Environment Understanding} module receives and processes the 3D point cloud coming from the camera to recognize ground, terrain type and obstacles, it locates the exoskeleton in the environment, and it identifies the best foothold position for the next step; the \textit{Adaptive Gait Generation (AGG)} module represents the core of the system and it employs two Kernelized Movement Primitives (KMPs) models for encoding the gait trajectories of the swing foot and the support leg offline. Via-points and linear constraints based on the outcome of the \textit{Environment Understanding} are then introduced to adapt the learned trajectories at run-time during the exoskeleton usage.}
\label{fig:overview}
\end{figure*}

Several works are present in the literature that employed a DMP-based walking motion generator for LLEs. The majority of the approaches adopt the DMP to achieve a better human-robot integration in flat-ground walking \cite{chen_dynamic_2018, huang_hierarchical_2018, luo_trajectory_2022}. For example, in \cite{chen_dynamic_2018, luo_trajectory_2022}, DMPs are used to learn the user's gait kinematics during human-exoskeleton interactions and generate a more natural exoskeleton motion coordinated with the user's movement. Nevertheless, DMP-based solutions have also been proposed for environment-adaptive gait generation in stairs  \cite{zhang_study_2024, ma_gait_2018} and slope scenarios \cite{huang_adaptive_2020,zou_slope_2021}, also incorporating different learning techniques to identify the optimal parameters of the nonlinear component of motion, such as Locally Weighted Regression (LWR) \cite{zou_adaptive_2019}, Gaussian Mixture Regression (GMR) \cite{xu_dmp-based_2022} and Reinforcement Learning \cite{zhang_motion_2022,zhang_study_2024}. 
Despite these strengths, the generalization to new and unseen situations in DMPs is limited as they are not capable inherently to learn the stochastic variability of human gait from multiple demonstrations. To address this limitation, several extensions have been proposed. Stylistic DMPs (SDMPs) \cite{matsubara_learning_2010} introduce a style parameter to generate different motions from multiple reference trajectories and have been used to encode gait at different walking speeds \cite{ma_dynamic_2020}. Probabilistic Movement Primitives (ProMP) \cite{paraschos2013probabilistic} further improve adaptability by representing trajectories as probability distributions, and have been used in LLEs to learn a more accurate representation of human motion from multiple demonstrations \cite{wang_probabilistic_2023}. KMPs \cite{huang2019kernelized} represent a more recent extension of the probabilistic movement representation that does not require an explicit basis function definition as in ProMP, and allows a higher flexibility in trajectory adaptation. In \cite{zou_learning_2021}, KMPs have been used to generate personalized hip and knee joint trajectories for different walking speeds. KMPs also showed promising performance in adapting the gait to different terrains \cite{zou_terrain-adaptive_2023, yang2024adaptive}, such as flat-ground, slopes, and stairs through the definition of specific via-points/end-points. For example, \cite{yang2024adaptive} combines KMP learning with an artificial potential field for the run-time identification of the via-points. However, while the introduction of end-points has demonstrated promising performance in adapting to different step lengths and ground inclinations, under walking conditions that differ substantially from the initial demonstrations (e.g., stairs, obstacle crossing), the use of via-points for trajectory shaping may produce trajectories that fall outside the learned probabilistic distributions. This, in turn, risks compromising the statistical properties encoded in the model and generating gait trajectories that do not preserve the physiological characteristics of human locomotion. Additionally, state-of-the-art DMP and KMP-based approaches are limited to the imitation and adaptation of the swing leg motion, either in the task space \cite{zou_slope_2021} or in the joint space \cite{luo_trajectory_2022}; while the movement of the support leg is described by conventional dynamic models (e.g., linear inverted pendulum) with limited adaptability to different walking tasks and with no guarantees on the naturalness of the generated trajectories.

In summary, the current literature presents the following limitations: (i) most adaptive gait solutions are restricted to specific environments, lacking generalization to diverse real-world settings—complex scenarios such as stair descent and obstacle crossing are often neglected; (ii) integration with vision sensors is rarely addressed, as environmental geometry is typically assumed to be fully known, limiting robustness to real-world uncertainties; and (iii) existing imitation-based methods focus primarily on swing-foot adaptation, overlooking the support leg’s behavior. This work addresses these gaps by introducing an environment-aware AGG system that simultaneously learns swing and support leg motions from human demonstrations within a unified KMP framework. By formulating the AGG as a constrained optimization problem, the proposed method generates safe, statistically consistent trajectories across various indoor settings. Finally, its integration and validation on a commercial lower-limb exoskeleton equipped with an RGB-D camera demonstrate the robustness and practical effectiveness of the approach.

\begin{figure}[t]
    \centering  \includegraphics[width=0.65\columnwidth]{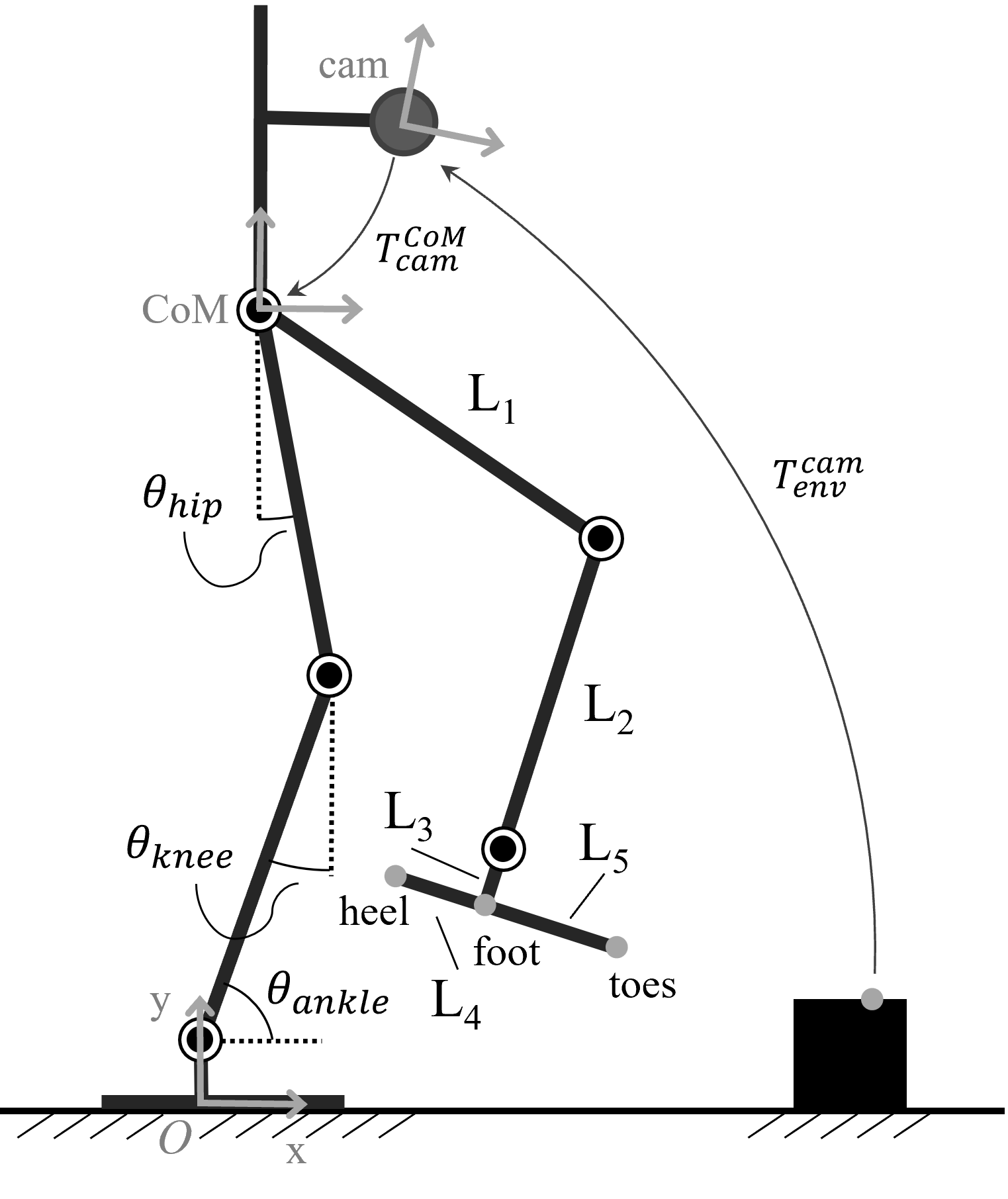}
        \caption{General kinematic model of the exoskeleton for the implementation of the proposed adaptive gait generation method.}
    \label{fig:kin_chain}
\end{figure}

\section{Methods}
\subsection{System Overview}
The full adaptive gait planning system proposed in this paper is summarized in Fig. \ref{fig:overview} and is composed of two main modules. The \textit{Environment Understanding} module is in charge of processing the visual input in order to identify the ground plane, classify the terrain type, and extract the main geometric characteristics of the environment, including the detection of possible obstacles. Additionally, it combines the camera information with the exoskeleton encoders to compute the 2D sagittal pose of the exoskeleton with respect to the environment (\textit{Exoskeleton Pose Estimation}), and identifies the next step foothold according to the type of terrain and the presence of obstacles (\textit{Foothold selection}). The \textit{Adaptive Gait Generation} (AGG) module is instead in charge of generating adaptive gait trajectories to drive the exoskeleton foot to the desired foothold position, avoiding collisions with the environment. To do this, two parallel KMP models are used for generating the swing foot trajectories in the task space and support leg angular trajectories in the joint space. In particular, the gait adaptation is achieved by combining the definition of via-points (e.g., to modify the main gait parameters such as step length and height) with constraints optimization for generating physiologically plausible and collision-free gait trajectories. Finally, the foot trajectories are converted into angular joint trajectories for the swing leg through inverse kinematics and sent to the low-level controller (e.g., PID motor control) of the exoskeleton for movement execution.

\subsection{Exoskeleton kinematic model}
Fig. \ref{fig:kin_chain} illustrates the kinematic chain of the generic lower limb exoskeleton that has been considered for the implementation of the proposed method. The anthropometric measurements of the exoskeleton are represented by the thigh length ($L_1$), the shin length ($L_2$), the foot height ($L_3$), the heel-to-foot length ($L_4$), and the foot-to-toes length ($L_5$), adapted for each user. Each leg of the exoskeleton is composed of three main joints: hip, knee, and ankle. In our exoskeleton platform, hips and knees are assumed to be active joints, e.g., controlled by brushless DC motors, while the ankles are assumed to be elastic passive joints with a resting angle equal to $\theta_{ankle}=90^{\circ}$ (e.g., when the foot is off the ground). This simplification is supported by the fact that it covers the majority of kinematic characteristics of both commercial and research exoskeleton platforms \cite{bettella2025scoping}. For each step, the base frame $\{O\}$ of the kinematic chain is located at the base of the support foot, while the end-effector is represented by the $foot$ position of the swing leg.

The kinematic chain is further enriched by additional links connecting an RGB-D camera to the frame of the exoskeleton's pelvis. The homogeneous transformation $T_{cam}^{CoM}$ between the camera frame and the exoskeleton's centre of mass (CoM) is assumed to be known, either by construction (e.g., from the CAD model) or through geometric estimations as in \cite{trombin2024environment}. Thus, given a point in the environment, captured by the camera ($T_{env}^{cam}$), the localization of the exoskeleton's base frame with respect to this point $T_{env}^{O}$ can be obtained by forward kinematics following the kinematic chain and knowing the angles $\theta_{hip}$, $\theta_{knee}$, and $\theta_{ankle}$ of each leg, e.g. measured by linear encoders at the exoskeleton joints. These computations are performed by the \textit{Exoskeleton Pose Estimation} module, and provided to the system for foothold selection and environment-aware gait generation.

\subsection{Environment Understanding}
\label{sec:environment_understanding}
The \textit{Environment Understanding} module aims at analysing the walking environment captured by an RGB-D camera. This is fundamental for three main tasks: (i) processing the input point cloud and identifying the main planes in the environment; (ii) identifying the terrain type (i.e., flat-ground, slopes, stairs) and estimating the main geometric characteristics of the environment; (iii) guiding the selection of the most appropriate position for the next foothold. 

\subsubsection{Point cloud processing and planes identification}
Our previous method \cite{trombin2024environment} assumed only indoor flat-ground terrain scenarios, easily identifiable with the RANSAC algorithm. In this section, we extend our approach to deal with different indoor walkable settings, such as stairs and slopes, and to be more robust to noise, outliers, and irregular surfaces characterizing real-world situations. 
In particular, the proposed method consists of a multi-plane segmentation strategy based on \emph{Maximum Likelihood Estimation Sample Consensus (MLESAC)} \cite{TORR2000138}. Unlike RANSAC, which minimizes the number of outliers, MLESAC evaluates the likelihood of each candidate model under the observed data distribution, thereby achieving greater robustness against noise and spurious measurements. This statistical grounding makes MLESAC particularly suitable for environments where planar structures are partially occluded or contaminated by clutter, conditions frequently encountered in locomotion scenarios. 
In practice, our method iteratively segments the input point cloud by combining normal estimation with likelihood-based plane fitting. The largest set of inliers corresponding to a planar surface is extracted and removed from the cloud at each iteration, allowing for the identification of multiple coexisting planes. This process naturally identifies elements of the environment, such as the ground, slopes, and stair treads for locomotion planning. Furthermore, a filtering process ensures that only geometrically relevant planes are kept: surfaces are taken only when their normal vector does not contain a predominant vertical component, i.e., \( |n_z| > 0.7 \), thereby making the assumption that only potentially traversable spaces are being considered. The ground plane is then selected as the closest plane to the subject’s feet, ensuring a physically consistent reference frame for motion analysis.   

\begin{figure}[t]
    \centering
     \includegraphics[width=\columnwidth]{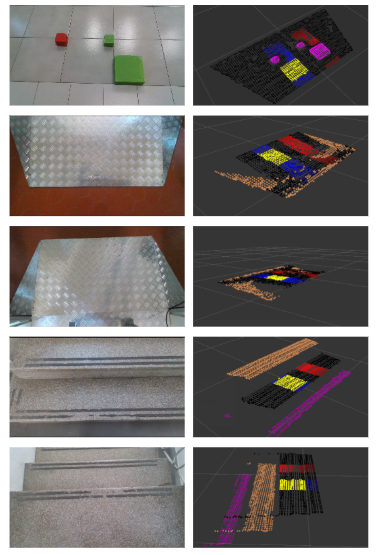}
    \caption{Examples of different terrains acquired with an RGB-D camera (left column) and the corresponding 3D point clouds processed with \textit{Environment Understanding} module (right column). The point clouds illustrate the segmentation of relevant areas for walking and the selected foothold position (yellow area).}
    \label{fig:pic_slopes_stairs}
\end{figure}

\subsubsection{Terrain classification and characterization}
After extracting the principal planes, a \emph{terrain classification} module assigns the scene to one of three categories based on plane orientation and spatial arrangement. A single horizontal plane indicates \emph{flat-ground}; multiple horizontal planes with vertical offset indicate \emph{stairs}; and the presence of a plane with an inclined normal indicates a \emph{slope}. The algorithm assumes that only one terrain type is visible per step. Depending on the detected terrain, specific geometric features are computed: stair height, tread length, and distance to the tread edge for stairs, and plane inclination for slopes. Residual non-planar points are then segmented using Euclidean clustering. Compact clusters near traversable planes are labeled as obstacles, whereas distant or sparse clusters are discarded as noise. For each obstacle, height, length, and distance from the support foot are estimated to construct a bounding box approximating its shape. A visualization of the outcome of the environment identification and segmentation is provided in Fig.~\ref{fig:pic_slopes_stairs}, which illustrates examples of RGB images of different terrains (on the left) alongside the corresponding segmented point clouds (on the right). The figure highlights the correct distinction between different planar surfaces corresponding to flat-ground, slopes and stair treads, as well as the correct clustering of multiple obstacles in the scene.

\subsubsection{Foothold selection}
The \textit{foothold selection} module is a modified version of the algorithm that we previously proposed in \cite{trombin2024environment}. The method identifies the optimal landing point for the swing foot from the processed 3D point cloud. 
In particular, a search area is defined in front of each foot—limited by the foot width and maximum step length represented in the blue and red areas of Fig.~\ref{fig:pic_slopes_stairs}. Each point $p$ in the search area is assigned a score that considered both the step cadence consistency and obstacle proximity:
\begin{equation}
    f(p) = f_G(p) \, f_O(p),
\end{equation}
where $f_G(p)$ is a Gaussian probability density function centered around the average step length and a variance equal to the physiological gait variability \cite{dambreville2015spinal}, while $f_O(p, O)$ is an occupancy function that penalizes proximity to obstacles or the intersections between different planes. We refer the reader to \cite{trombin2024environment} for a more detailed description of the functions. Then, a moving window $S_w$, matching the foot dimensions (i.e., $L_4$ + $L_5$), is shifted along the search area to compute a local average score. The window with the maximum score,
\begin{equation}
    F_{\text{next}} = \max_{w} S_w,
\end{equation}
is selected as the next foothold and used to compute the final foot position $(x_{foot}^{final},y_{foot}^{final})$ for the AGG. The selection of the foothold position for the next step is represented by the yellow areas in Fig.~\ref{fig:pic_slopes_stairs}.

\subsection{Learning gait from human demonstrations}
\label{sec:hybrid_kmp}
In order to have a probabilistic representation of gait, in this work KMP model is employed to learn gait trajectories from human demonstrations. Kernelized Movement Primitives (KMP) \cite{huang2019kernelized} is a learning-based approach that provides a nonparametric, kernel-based formulation for imitating demonstrated trajectories (e.g., from a human gait database) while allowing principled adaptation to new conditions. 
Let us consider a set of demonstrated training data denoted by $\{ \{{s}_{n,h}, \boldsymbol{\xi}_{n,h} \}_{n=1}^{N} \}_{h=1}^{H}$ where \textit{H} and \textit{N} represent the number of demonstrations and the number of trajectory points per demonstration, respectively. $\boldsymbol{\xi}_{n,h} \in \mathbb{R}^\mathcal{O}$ is the gait trajectory and ${s}_{n,h} \in [0,1]$ is a phase variable linked to the time $t$ as $t=s_{n,h}T$ so that the model can generalize for any step duration $T$. 
Differently from other KMP-based controller, in this work we adopt a hybrid approach that learns the human gait both in the task space and in the joint space. To do so, we employ two KMP models: one model learns the trajectory of the swing foot in the task space, thus $\boldsymbol{\xi}_{n,h}^{foot}=(x_{foot}({s}_{n,h}), y_{foot}({s}_{n,h}))$ represents the foot position in the sagittal plane. The second model is instead introduced to model the dynamic behavior of the centre of mass by encoding the movement of the support leg in the joint space; therefore, $\boldsymbol{\xi}_{n,h}^{supp}=(\theta_{hip}({s}_{n,h}),\theta_{knee}({s}_{n,h}))$ is the joint configuration of the support leg described by its hip and knee joints' angle. Thus, for both KMP models the output dimension is $\mathcal{O}=2$. To avoid repetitions, all the following formulations are applied to both the swing foot model and the support leg model, when not stated otherwise.

For each KMP model, a Gaussian Mixture Model (GMM) is employed to learn the probabilistic distribution of the demonstrations \cite{michieletto2015gmm}. The GMM estimates the following joint probability distribution:
\begin{equation}
    \begin{bmatrix}
        \mathbf{s} \\ 
        \boldsymbol{\xi}
    \end{bmatrix}
    \sim \sum_{c=1}^{C} \pi_c \mathcal{N}(\boldsymbol{\mu}_c, \boldsymbol{\Sigma}_c)
\end{equation}
where ${\pi}_c$, $\boldsymbol{\mu}_c$, and $\boldsymbol{\Sigma}_c$ represent the prior probability, mean, and covariance of the $c$-th Gaussian component, respectively, and $C$ indicates the number of Gaussian components. After calibrating the GMM on the demonstrated training dataset, a probabilistic reference trajectory $\{\hat{\boldsymbol{\xi}}_n\}_{n=1}^N$ is retrieved using Gaussian Mixture Regression (GMR), where each trajectory point $\hat{\boldsymbol{\xi}}_n$ is described by a conditional probability distribution with mean $\hat{\boldsymbol{\mu}}_n$ and covariance $\hat{\boldsymbol{\Sigma}}_n$, i.e., $\hat{\boldsymbol{\xi}}_n | {s}_n \sim \mathcal{N}(\hat{\boldsymbol{\mu}}_n, \hat{\boldsymbol{\Sigma}}_n)$. This information is saved in the \textit{reference database} $\mathbf{D} = \{{s}_n, \hat{\boldsymbol{\mu}}_n, \hat{\boldsymbol{\Sigma}}_n\}_{n=1}^{N}$, containing the probability distribution of each point of the reference trajectory. 

To derive the KMP solution, the trajectory is represented as a parametric function expressed as a weighted combination of basis functions:

\begin{equation}
\label{eq:traj_out}
\boldsymbol{\xi}(s) = \boldsymbol{\Theta}(s)^\top \mathbf{w}
\end{equation}

where $\boldsymbol{\Theta}(s) \in \mathbb{R}^{\mathcal{B} \mathcal{O} \times \mathcal{O} }$ is a matrix of $\mathcal{B}$-dimensional basis functions, and $\mathbf{w} \sim  \mathcal{N}(\boldsymbol{\mu}_w, \boldsymbol{\Sigma}_w) \in \mathbb{R}^{\mathcal{B} \mathcal{O}}$ is a vector of normally distributed weights with mean $\boldsymbol{\mu}_w$ and variance $\boldsymbol{\Sigma}_w$ defined as \cite{huang2019kernelized}
\begin{equation}
\begin{aligned}
    &\boldsymbol{\mu}_w = \boldsymbol{\Phi}(\boldsymbol{\Phi}^\top\boldsymbol{\Phi} + \lambda\boldsymbol{\Sigma})^{-1}\boldsymbol{\mu}\\
    &\boldsymbol{\Sigma}_w = N(\boldsymbol{\Phi}\boldsymbol{\Sigma}^{-1}\boldsymbol{\Phi}^\top + \lambda_c\boldsymbol{I})^{-1}
\end{aligned}
\end{equation}
where
\begin{equation}
\begin{aligned}
    &\boldsymbol{\Phi} = [\boldsymbol{\Theta}(s_1) \boldsymbol{\Theta}(s_2) \dots \boldsymbol{\Theta}(s_{N})] \\
    &\hat{\boldsymbol{\Sigma}} = \text{blockdiag}(\hat{\boldsymbol{\Sigma}}_1, \hat{\boldsymbol{\Sigma}}_2, \dots, \hat{\boldsymbol{\Sigma}}_N), \\
    &\hat{\boldsymbol{\mu}} = [\hat{\boldsymbol{\mu}}_1^\top \ \hat{\boldsymbol{\mu}}_2^\top \ \dots \ \hat{\boldsymbol{\mu}}_N^\top]^\top
\end{aligned}
\label{eq:phi_mu_sigma}
\end{equation}
and $\lambda>0$, $\lambda_c>0$ are regularization parameters. 

\begin{figure*}[t]
    \centering
    \includegraphics[width=\textwidth]{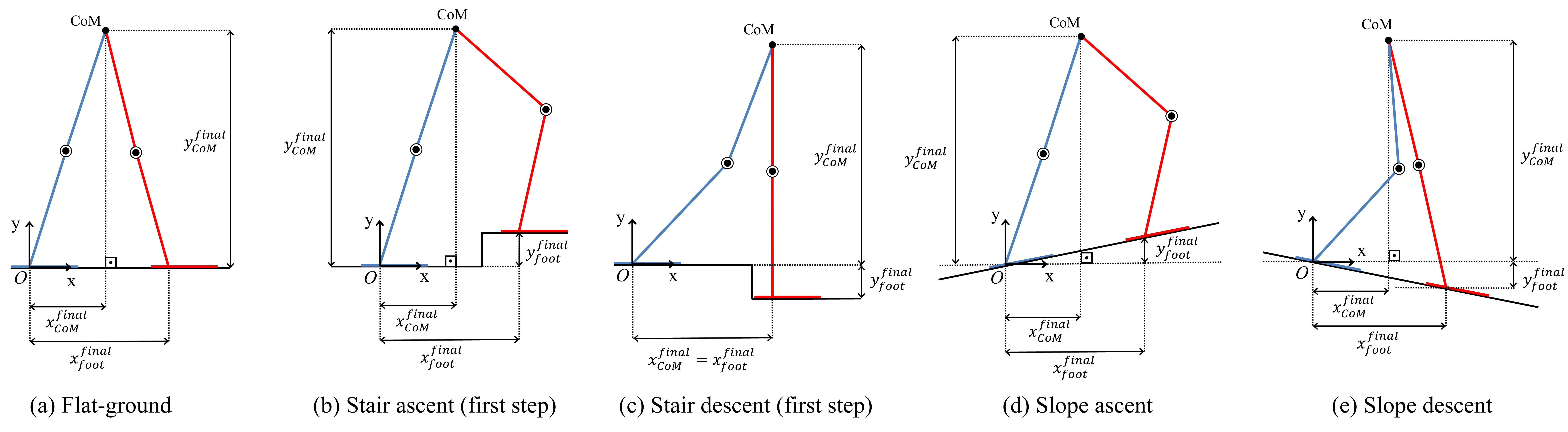}
    \caption{Final exoskeleton configuration for local frame computation in different terrains.}
    \label{fig:com}
\end{figure*}

To avoid an explicit formulation of the basis functions, KMP uses kernel functions to model the trajectory distribution, linked to the basis function matrix as follows
\begin{equation}
    \mathbf{k}({s}_i, {s}_j) = \boldsymbol{\Theta}(s_{i})^\top \boldsymbol{\Theta}(s_{j}) = k ({s}_i, {s}_j) \mathbf{I}_{\mathcal{O}}
    \label{eq:kernel_matrix}
\end{equation}
where $\mathbf{I}_{\mathcal{O}}$ is the $\mathcal{O}$-dimensional identity matrix, and forming the kernel matrix $\mathbf{K}$ denoted as
\begin{equation}
    \mathbf{K} = 
    \begin{bmatrix}
        \mathbf{k}({s}_1, {s}_1) & \mathbf{k}({s}_1, {s}_2) & \cdots & \mathbf{k}({s}_1, {s}_N) \\
        \mathbf{k}({s}_2, {s}_1) & \mathbf{k}({s}_2, {s}_2) & \cdots & \mathbf{k}({s}_2, {s}_N) \\
        \vdots & \vdots & \ddots & \vdots \\
        \mathbf{k}({s}_N, {s}_1) & \mathbf{k}({s}_N, {s}_2) & \cdots & \mathbf{k}({s}_N, {s}_N)
    \end{bmatrix}
    \label{eq:kernel_matrix}
\end{equation}
In this work, the Gaussian kernel function is chosen:
\begin{equation}
    k({s}_i, {s}_j) = \exp\left(-l \|{s}_i - {s}_j\|^2 \right)
    \label{eq:kernel_function}
\end{equation}
where $l>0$ is a parameter that influences the decay of the kernel values and ensure appropriate function smoothness and model generalization.

\subsection{Adaptive gait trajectory generation}
\label{sec:localkmp}
Once the KMP models are calibrated offline, we can use these models to generate new gait trajectories during exoskeleton control. This process corresponds to the generation of the output trajectory $\tilde{\boldsymbol{\xi}}(\mathbf{s}^*)$ for a set of $N$ query points $\mathbf{s}^*=\{{s}^*_n\}_{n=1}^N$ that preserves the probabilistic distribution of the reference trajectory. 
In our AGG method, we need to adapt the exoskeleton's step to a wide range of movements necessary to walk through different types of terrains and unconstrained settings (e.g., in the presence of obstacles). This could lead to the generation of trajectories that can considerably differ from the nominal reference trajectory. To deal with this challenge, a local KMP formulation \cite{huang2019kernelized} is introduced. This variant of the KMP formulation allows to extract local movement patterns by computing multiple new trajectories in different reference frames, which are then combined to obtain a global trajectory, ensuring a better coherence with the reference trajectory even when the generated trajectory deviates significantly from the demonstrations used for KMP training (e.g., for the introduction of new via-points). 

To do so, during trajectory generation we define $P$ local frames, denoted as $\{\mathbf{A}^{(p)}, \mathbf{b}^{(p)}\}_{p=1}^P$, where each frame $\{p\}$ is characterized by a rotation matrix $\mathbf{A}^{(p)}$ and a translation vector $\mathbf{b}^{(p)}$ relative to the base frame $\{O\}$. In our case, we locate the base frame $\{O\}$ at the support foot, and we consider two local frames (i.e., $P=2$), one at the start of the step trajectory and one at the end of the step trajectory, for both the swing foot KMP model and the support leg KMP model. After defining the local frames, the demonstrations used for training the KMP models are transformed into each local frame $\{p\}$, yielding a new set of trajectory points $\{\{{s}_{n, h}^{(p)}, \boldsymbol{\xi}_{n, h}^{(p)}\}_{n=1}^N\}_{h=1}^H$  within each local frame, with the following transformation
\begin{equation}
\label{eq:local_project}
    \begin{bmatrix}
        {s}_{n, h}^{(p)} \\
        \boldsymbol{\xi}_{n, h}^{(p)}
    \end{bmatrix}
    =
    \begin{bmatrix}
        {A}_s^{(p)} & \mathbf{0} \\
        \mathbf{0} & \mathbf{A}_\xi^{(p)}
    \end{bmatrix}^{-1}
    \left(
    \begin{bmatrix}
        {s}_{n, h} \\
        \boldsymbol{\xi}_{n, h}
    \end{bmatrix}
    -
    \begin{bmatrix}
        {b}_s^{(p)} \\
        \mathbf{b}_\xi^{(p)}
    \end{bmatrix}
    \right).
\end{equation}
Similarly, query inputs $\mathbf{s}^*$ in the base frame $\{O\}$ should also be projected into local frames using the transformation of Eq. \eqref{eq:local_project}, yielding local query inputs $\{\mathbf{s}^{*(p)}\}_{p=1}^P$. As before, we can use GMM and GMR to produce a \textit{local reference database} $\mathbf{D}^{(p)} = \{{s}^{(p)}_n, \hat{\boldsymbol{\mu}}_n^{(p)}, \hat{\boldsymbol{\Sigma}}_n^{(p)}\}_{n=1}^N$ for each local frame $\{p\}$. With this formulation, the aim is to predict a local trajectory $\tilde{\boldsymbol{\xi}}^{(p)}(\mathbf{s}^{*(p)}) \sim \mathcal{N}(\boldsymbol{\mu}^{*(p)},\boldsymbol{\Sigma}^{*(p)})$ within each frame $\{p\}$, with local mean $\boldsymbol{\mu}^{*(p)}$ and covariance $\boldsymbol{\Sigma}^{*(p)}$. The final output trajectory $\tilde{\boldsymbol{\xi}}(\mathbf{s}^*)$ can be then retrieved in the base frame $\{O\}$ as
\begin{equation}
\label{eq:local_to_global}
    \tilde{\boldsymbol{\xi}}(\mathbf{s}^*) = \left(\sum_{p=1}^P \tilde{\boldsymbol{\Sigma}}_p^{-1} \right)^{-1} \sum_{p=1}^P \tilde{\boldsymbol{\Sigma}}_p^{-1} \tilde{\boldsymbol{\mu}}_p
\end{equation}
where
\begin{equation}
    \begin{aligned}
        \tilde{\boldsymbol{\mu}}_p &=\mathbf{A}_\xi^{(p)} \boldsymbol{\mu}^{*(p)} + \mathbf{b}_\xi^{(p)} \\
        \tilde{\boldsymbol{\Sigma}}_p &= \mathbf{A}_\xi^{(p)} \boldsymbol{\Sigma}^{*(p)} \mathbf{A}_\xi^{(p)\top}
    \end{aligned}
\end{equation}

In our AGG, the following assumptions are made that simplify the identification of the local frames: first, knowing that the phase is time-related and does not depend on the local frame, the phase rotation and translation is simplified to ${A}_s^{(p)}=1$ and ${b}_s^{(p)}=0$, thus $s_{n,h}={s}_{n, h}^{(p)}$ and $\mathbf{s}^*=\mathbf{s}^{*(p)}$. Second, we assume that the local frame $\{p\}$ has the same orientation as the base frame $\{O\}$, thus $\mathbf{A}_\xi^{(p)}=\boldsymbol{I}$. Thus, the transformation to the local frames is reduced to the identification of the vectors $\mathbf{b}^{(0)}_{\xi,{foot}} = (x_{foot}^{init},y_{foot}^{init})$ and $\mathbf{b}^{(1)}_{\xi,foot} = (x_{foot}^{final},y_{foot}^{final})$ for the swing foot KMP model, and the vectors $\mathbf{b}^{(0)}_{\xi,supp} = (\theta_{hip}^{init},\theta_{knee}^{init})$ and $\mathbf{b}^{(1)}_{\xi,supp} = (\theta_{hip}^{final},\theta_{knee}^{final})$ for the support leg KMP model, that depends on the actual exoskeleton configuration and specific terrain type. In particular, the initial frames $\mathbf{b}^{(0)}_{\xi,foot}$ and $\mathbf{b}^{(0)}_{\xi,supp}$ are provided by the \textit{Exoskeleton Pose Estimation} at run-time. The final frame $\mathbf{b}^{(1)}_{\xi,foot}$ of the final swing foot position is instead provided by the \textit{Foothold selection}. For what concerns instead the final frame for the support leg model $\mathbf{b}^{(1)}_{\xi,supp}$, it is computed based on the desired final position of the exoskeleton's CoM $(x_{CoM}^{final},y_{CoM}^{final})$ according to the type of terrain, as shown in Fig. \ref{fig:com}. Once this position is defined, the corresponding support leg joint angles $(\theta_{hip}^{final},\theta_{knee}^{final})$ are computed via inverse kinematics and used as final frame $\mathbf{b}^{(1)}_{\xi,supp}$ for the support leg model.  
In the following, a more detailed description of the final exoskeleton configuration for each terrain is provided.

\subsubsection{Flat-ground}
in the case of normal walking on a flat-ground surface, the final horizontal position of the hip is set to the midpoint between the feet:
\begin{equation}
    x_{com}^{final} = \frac{x_{foot}^{final}}{2}
    \label{eq:hor_hip_pos}
\end{equation}
The vertical position of the hip $y_{com}^{final}$ is then calculated geometrically, assuming that both knees are fully extended as in Fig. \ref{fig:com}a.

\subsubsection{Stairs ascent and descent} 
for stair negotiation, we divided the problem into two scenarios: stair ascent and stair descent. In stair ascent, during the first step onto the stair tread (Fig. \ref{fig:com}b), the horizontal hip position is also at the midpoint between the feet as in Eq. \eqref{eq:hor_hip_pos}, while the vertical position is computed geometrically under the assumption that only the support leg's knee is fully extended at the end of the movement. During stair descent, in the first step off the stair tread (Fig. \ref{fig:com}c), the final CoM position is set above the swing foot (i.e., ${x}_{com}^{final}={x}_{foot}^{final}$) with the swing leg fully extended at the end of the movement. This configuration ensures, on the one hand, a high degree of stability by aligning the system’s centre of mass with its centre of pressure. On the other hand, it allows the load to be relieved from the rear leg, thereby facilitating the execution of the subsequent step. The second step in both ascent and descent stairs condition is constrained so that the swing foot position coincides with the support foot position in a straight standing configuration, thus
\begin{equation}
\begin{aligned}
    x_{com}^{final}&=x_{foot}^{final}=y_{foot}^{final}=0\\
    y_{com}^{final}&=l_{leg}
\end{aligned}
\end{equation}
where $l_{leg}=L_1+L_2+L_3$ is the leg length. The two steps can then be repeated to ascend or descend the staircase one tread at a time.

\subsubsection{Slope ascent and descent}
Walking on inclined surfaces also requires distinct strategies for ascending (Fig. \ref{fig:com}d) and descending (Fig. \ref{fig:com}e). Similarly to the flat-ground walking condition, the final CoM horizontal position is set to the midpoint between the feet following Eq. \eqref{eq:hor_hip_pos}, while the vertical $y_{com}^{final}$ position is computed assuming a fully extended stance leg for slope ascending, and a fully extended swing leg for slope descending. With this strategy, the CoM is always shifted towards the high-ground side of the slope, ensuring stability and user comfort.

With the above-mentioned approach, our method is able to generate different local trajectories using the pre-trained KMP models for different terrains. The final step is the adaptation of the gait trajectories to the desired step parameters (e.g., step length, step height) and to the geometric characteristics of the environment provided by the \textit{Environment Understanding} module described in Section \ref{sec:environment_understanding}. To do so, we extend the formulation by defining the AGG problem as a linearly constrained optimization problem with modifiable via-points.

\subsection{Step parameters modification with via-points}
To tailor the generated gait trajectory to the desired step characteristics (e.g., step length, step height), we exploit the introduction of additional via-points that the resulting trajectory is constrained to pass through. Formally, the process consists in defining the new desired via-points as $\{\bar{{s}}_j, \bar{\boldsymbol{\xi}}_j\}_{j=1}^J$ associated with conditional probability distributions $\bar{\boldsymbol{\xi}}_j | \bar{{s}}_j \sim \mathcal{N}(\bar{\boldsymbol{\mu}}_j, \bar{\boldsymbol{\Sigma}}_j)$, where $\bar{\boldsymbol{\mu}}_j$ is the position of the via-point, $\bar{\boldsymbol{\Sigma}}_j$ determines how strong the generated trajectory is forced to pass through the point, and $J$ is the number of new points. In particular, we set the covariance as $\bar{\boldsymbol{\Sigma}}_j = 10^{-7}\hspace{1mm}\mathbf{I}_2$, where $\mathbf{I}_2$ is the two-dimensional identity matrix.  
Once the new via-points are defined, each \textit{local reference database} $\mathbf{D}^{(p)}$ is updated at run-time by adding the new points $\{\bar{{s}}_j, \bar{\boldsymbol{\xi}}_j^{(p)}\}_{j=1}^J$ projected in the frame $\{p\}$, checking that the distance between the new via-point is sufficiently far apart from the nearest point in $\mathbf{D}^{(p)}$, called ${s}_d$, i.e., $\left\|\bar{{s}}_j-{s}_d\right\|<\zeta$ with $\zeta>0$ being a threshold. Whenever this condition is not met, the data point $\{{s}_d, \hat{\boldsymbol{\mu}}_d^{(p)},\hat{\boldsymbol{\Sigma}}_d^{(p)}\}$ is removed from the database $\mathbf{D}^{(p)}$ before adding the new via-point:
\begin{equation}
\label{eq:update_via}
    \begin{aligned}
        \mathbf{D}^{(p)} \leftarrow &\{ \mathbf{D}^{(p)} \setminus \{{s}_d, \hat{\boldsymbol{\mu}}_d^{(p)}, \hat{\boldsymbol{\Sigma}}_d^{(p)}\} \} 
        \\ &\cup\{\bar{{s}}_j, \bar{\boldsymbol{\mu}}_j^{(p)},\hat{\boldsymbol{\Sigma}}_j^{(p)}\}, \text{if } \left\| \bar{{s}}_j - {s}_d \right\| < \zeta, \\
        \mathbf{D}^{(p)} \leftarrow &\mathbf{D}^{(p)} \cup \{\bar{{s}}_j, \bar{\boldsymbol{\mu}}_j^{(p)},\hat{\boldsymbol{\Sigma}}_j^{(p)}\}, \text{otherwise.}
    \end{aligned}
\end{equation}

In particular, for each step, we add the initial and final exoskeleton positions as starting and ending points of the local reference trajectories. Thus, the database points related to ${s}_1$ and ${s}_N$ are updated with the expected values $\hat{\boldsymbol{\mu}}_{1}^{(p)}$ and $\hat{\boldsymbol{\mu}}_{N}^{(p)}$ based on the outcome of the \textit{Exoskeleton Pose Estimation} and \textit{Foothold selection}, for both the swing foot model and the support leg model. This step is fundamental for the correct computation of the local frames, as explained in Section \ref{sec:localkmp}, and to impose a desired step length of the exoskeleton gait.

On the other hand, to modify the desired step height, we introduce an additional via-point in the swing foot model. In particular, the database point corresponding to the peak of the reference foot trajectory ${s}_m \in \mathbf{D}^{(p)}$ with $\hat{{\mu}}_{h,y}^{(p)} = \max_n (\hat{{\mu}}_{n,y}^{(p)}), \hspace{2mm} n \in \{1,...,N\}$, is replaced with the new via-point $\bar{{s}}_j = {s}_m$,  $\bar{\boldsymbol{\mu}}_{j}^{(p)} = \{\hat{{\mu}}_{h,x}^{(p)}, stepHeight^{(p)}\}$, where $stepHeight^{(p)}$ is the desired step height with respect to the support foot position projected in the local frame $\{p\}$. 

Thanks to this formulation, the gait generation model can be adapted to accommodate different types of step and customize the gait parameters to the user's preference or the kinematic limits of the exoskeleton platform.


\subsection{Linearly Constrained KMP for obstacle avoidance}
\label{sec:lc_kmp}
In addition to customizing the primary gait parameters, a high degree of adaptability in the shape of the generated trajectories is required to effectively handle potential obstacles or hazardous regions in the environment (e.g., the edge of a stair step), while preserving the statistical characteristics of the reference trajectories learned by the KMP models. To do so, we introduce in our system an extension of the KMP framework known as Linearly Constrained KMP (LC-KMP) \cite{huang_linearly_2020}. With LC-KMP, we formulate the generation of an arbitrary gait trajectory as a constrained optimization problem that incorporates linear equality and inequality constraints in the querying of the gait trajectories. Formally, the problem can be expressed as follows: 
\begin{align}
    \mathbf{w}^{*(p)} = \operatorname*{argmax}_{\mathbf{w}} &\sum_{n=1}^N \mathcal{P}\big(\boldsymbol{\xi}^{(p)}({s}_n) \big| \hat{\boldsymbol{\mu}}_n^{(p)}, \hat{\boldsymbol{\Sigma}}_n^{(p)}\big) \\
    &\mathbf{g}_{n,1}^{\top}\boldsymbol{\xi}^{(p)}({s}_n) \geq c_{n,1} \nonumber \\
    &\mathbf{g}_{n,2}^{\top}\boldsymbol{\xi}^{(p)}({s}_n) \geq c_{n,2} \nonumber \\
    \mathbf{s.t.} \quad &\qquad \vdots \quad \qquad \qquad ,\forall n \in \{1, 2, \dots, N\}  \nonumber \\
    &\mathbf{g}_{n,F}^{\top}\boldsymbol{\xi}^{(p)}({s}_n) \geq c_{n,F} \nonumber \\
\end{align}
where $\mathbf{w}^{*(p)}$ is the optimal weight vector to produce the local trajectory according to Eq. \eqref{eq:traj_out}, $\mathbf{g}_{n,f} \in \mathbb{R}^{\mathcal{O}}$ and $c_{n,f} \in \mathbb{R}$ characterize the $f$-th linear constraint on the local trajectory point $\boldsymbol{\xi}^{(p)}({s}_n)$, and $F$ denotes the total number of constraints.

By introducing Lagrange multipliers, the problem becomes a quadratic optimization problem with linear constraints, which can be solved by the classical \textit{quadratic programming} to find the optimal Lagrange multipliers $\boldsymbol{\alpha}^*$ through maximizing
\begin{equation}
\label{eq:lagrange_multi}
    \begin{aligned}
        \tilde{L}(\boldsymbol{\alpha}) = \ &\boldsymbol{\alpha}^\top \mathbf{B}_1 \boldsymbol{\alpha} + \mathbf{B}_2 \boldsymbol{\alpha}, \\
        \text{s.t.} \quad &\boldsymbol{\alpha} \geq 0.
    \end{aligned}
\end{equation}
where
\begin{equation}
    \begin{aligned}
        &\mathbf{B}_1 = \overline{\mathbf{G}}^\top \hat{\mathbf{\Sigma}}^{(p)} \mathbf{A} \hat{\mathbf{\Sigma}}^{(p)} \overline{\mathbf{G}}, \\ 
        &\mathbf{B}_2 = 2 \hat{\boldsymbol{\mu}}^{(p)\top} \mathbf{A} \hat{\mathbf{\Sigma}}^{(p)} \overline{\mathbf{G}} + \overline{\mathbf{C}}^\top \\
        &\mathbf{A} = -\frac{1}{2} (\mathbf{K} + \lambda \hat{\mathbf{\Sigma}}^{(p)})^{-1} \big(\mathbf{K} \hat{\mathbf{\Sigma}}^{(p)-1} \mathbf{K} + \lambda \mathbf{K}\big) (\mathbf{K} + \lambda \hat{\mathbf{\Sigma}}^{(p)})^{-1}
    \end{aligned}
\end{equation}
where $\hat{\boldsymbol{\mu}}^{(p)}$ and $\hat{\boldsymbol{\Sigma}}^{(p)}$ are the local frame projections of the reference mean and variance matrices in Eq. \eqref{eq:phi_mu_sigma}, and 
\begin{equation}
\label{eq:matrices_G_C}
    \begin{aligned}
    &\mathbf{G}_n = [\mathbf{g}_{n,1} \ \mathbf{g}_{n,2} \ \dots \ \mathbf{g}_{n,F}], \quad \forall n \in \{1, 2, \dots, N\}\\
    &\overline{\mathbf{G}} = \text{blockdiag}(\mathbf{G}_1, \ \mathbf{G}_2,\ \dots, \ \mathbf{G}_N) \\
    &\mathbf{C}_n = [c_{n,1} \ c_{n,2} \ \dots \ c_{n,F}]^\top, \quad \forall n \in \{1, 2, \dots, N\} \\
    &\overline{\mathbf{C}} = [\mathbf{C}_1^\top \ \mathbf{C}_2^\top \ \dots \ \mathbf{C}_N^\top]^\top.
    \end{aligned}
\end{equation}

Now, given the optimal $\boldsymbol{\alpha}^*$ and a set of input query points $\mathbf{s}^*=\{s^*_n\}_{n=1}^N$, the predicted output trajectory for each local frame can be found as 
\begin{equation}
    \label{eq:predict_LC_KMP}
    \tilde{\boldsymbol{\xi}}^{(p)}({s}_n^*) = \mathbf{k}^* (\mathbf{K} + \lambda \hat{\boldsymbol{\Sigma}}^{(p)})^{-1}(\hat{\boldsymbol{\mu}}^{(p)} + \hat{\boldsymbol{\Sigma}}^{(p)} \overline{\mathbf{G}} \boldsymbol{\alpha}^*)
\end{equation}
where $\mathbf{K}$ is the kernel matrix in Eq. \eqref{eq:kernel_matrix} and
\begin{equation}
    \mathbf{k}^* = [\mathbf{k}({s}_n^*, {s}_1) \ \mathbf{k}({s}_n^*, {s}_2) \ \dots \ \mathbf{k}({s}_n^*, {s}_N)]
    \label{eq:kernel_star}
\end{equation}
which is obtained from the minimization of the Kullback–Leibler (KL) divergence between the output trajectory distribution and the reference trajectory distribution \cite{huang2019kernelized}. All the computed local trajectories can be finally combined according to Eq. \eqref{eq:local_to_global} to generate the linearly constrained gait trajectory in the global frame.

In our AGG system, we considered two sets of constraints for both the swing foot and support leg KMP models: constraints associated with \textit{Trajectory Boundaries} and constraints associated with \textit{Obstacle Avoidance}. 

\subsubsection{Trajectory Boundaries}
This first set of constraints is introduced to ensure that the generated trajectories are feasible and respect the kinematic limits of the exoskeleton platform. In particular, we introduce four constraints applied to all the $N$ points of the trajectory defined as follows
\begin{equation}
\label{eq:sup_traj_const}
\begin{aligned}
    &\mathbf{g}_{n,1}^T = [1,0] \hspace{6mm} c_{n,1}=\theta^{hip}_{low}  \\ 
    &\mathbf{g}_{n,2}^T = [-1,0] \hspace{3mm} c_{n,2}=\theta^{hip}_{up}  \\ 
    &\mathbf{g}_{n,3}^T = [0,1] \hspace{6mm} c_{n,3}=\theta^{knee}_{low} \\ 
    &\mathbf{g}_{n,4}^T = [0,- 1] \hspace{3mm} c_{n,4}=\theta^{knee}_{up} \\ 
    &\forall n \in \{1, 2, \dots, N\} 
\end{aligned}
\end{equation}
for the support leg KMP model, and
\begin{equation}
\label{eq:swg_traj_const}
\begin{aligned}
    &\mathbf{g}_{n,1}^T = [1,0] \hspace{6mm} c_{n,1}=x_{low}  \\ 
    &\mathbf{g}_{n,2}^T = [-1,0] \hspace{3mm} c_{n,2}=x_{up}  \\ 
    &\mathbf{g}_{n,3}^T = [0,1] \hspace{6mm} c_{n,3}=y_{low}(n) \\ 
    &\mathbf{g}_{n,4}^T = [0,- 1] \hspace{3mm} c_{n,4}=y_{up} \\ 
    &\forall n \in \{1, 2, \dots, N\} 
\end{aligned}
\end{equation}
for the swing foot KMP model. The values $(\theta^{hip}_{low},\theta^{hip}_{up},\theta^{knee}_{low},\theta^{knee}_{up})$ and $(x_{low},x_{up},y_{low},y_{up})$ represent lower and upper boundaries for the generated trajectories in the joint space and in the task space, respectively.

For the support leg model, the hip angle is constrained between its initial and final values according to the type of terrain, computed as explained in Section \ref{sec:localkmp}, thus $\theta^{hip}_{low}= \theta_{hip}({s}_N)$, $\theta^{hip}_{up}= \theta_{hip}({s}_1)$. This constraint ensures a monotonic trend of the hip joint, respecting its physiological behavior in natural human walking \cite{boo_comprehensive_2025}. On the other hand, the knee angle is only upper-bounded by setting, $\theta^{knee}_{low}= -\infty$ and $\theta^{knee}_{up}= 0$, avoiding possibly dangerous over-extensions of the knee joint.

For the swing foot KMP model, the x-axis boundaries are simply determined by the starting and ending foothold positions, thus $x_{low}= x_{foot}({s}_1)$, $x_{up}= x_{foot}({s}_N)$, ensuring a forward progression of the swing foot. The upper bound of the y-axis is set as $y_{up}= maxStepHeight$, and depends on the mechanical characteristics of the exoskeleton platform. Different from the previous constraints, the lower bound of the y-axis is defined as a function of the trajectory phase index $n$, and it is adapted according to the type of walking terrain to avoid generating trajectories colliding with the ground plane. For flat-ground terrain, the boundary is simply set for each trajectory point as $y^{flat}_{low}= min({y}_{foot}({s}_1), {y}_{foot}({s}_N))$. For the slope ascent and descent terrains, the inclination of the ground is considered with the following constraint:
\begin{equation} 
y^{slope}_{low}(n) = \frac{{y}_{foot}({s}_N)-{y}_{foot}({s}_1)}{{x}_{foot}({s}_N)-{x}_{foot}({s}_1)} \hat{\mu}_{n,x}
\end{equation}
where $\hat{\mu}_{n,x}$ represents the x-coordinate of the reference trajectory at index $n$.

In the case of stairs, we can describe the lower boundary $y^{stair}_{low}$ as a piece-wise function that depends on the position of the edge of the stair tread $x_r$ as follows
\begin{equation}
\begin{aligned}
y^{stair}_{low}(n)  = \left\{ \begin{array}{cl}
y_{foot}(s_1) & : \ n < n_r \\
y_{foot}(s_N) & : \ n \geq n_r
\end{array} \right.\\ 
n_r = \min_n (||\hat{\mu}_{n,x} - x_{r}||), \hspace{3mm} n \in \{1,...N\}
\end{aligned}
\end{equation}

\subsubsection{Obstacle Avoidance}
This second set of constraints is applied to the swing foot KMP model only, to ensure the generation of collision-free trajectories for each point of the foot link. In this case, for the identification of the linear constraints the following considerations are made: (i) the full CoM trajectory is known by applying the forward kinematics to the local support leg angular trajectories $\tilde{\boldsymbol{\xi}}^{(p)}_{supp}(\mathbf{s}^*)$; (ii) the obstacle's surface can be discretized in $M$ equidistant points. 

\begin{figure}[t]
    \centering  \includegraphics[width=\columnwidth]{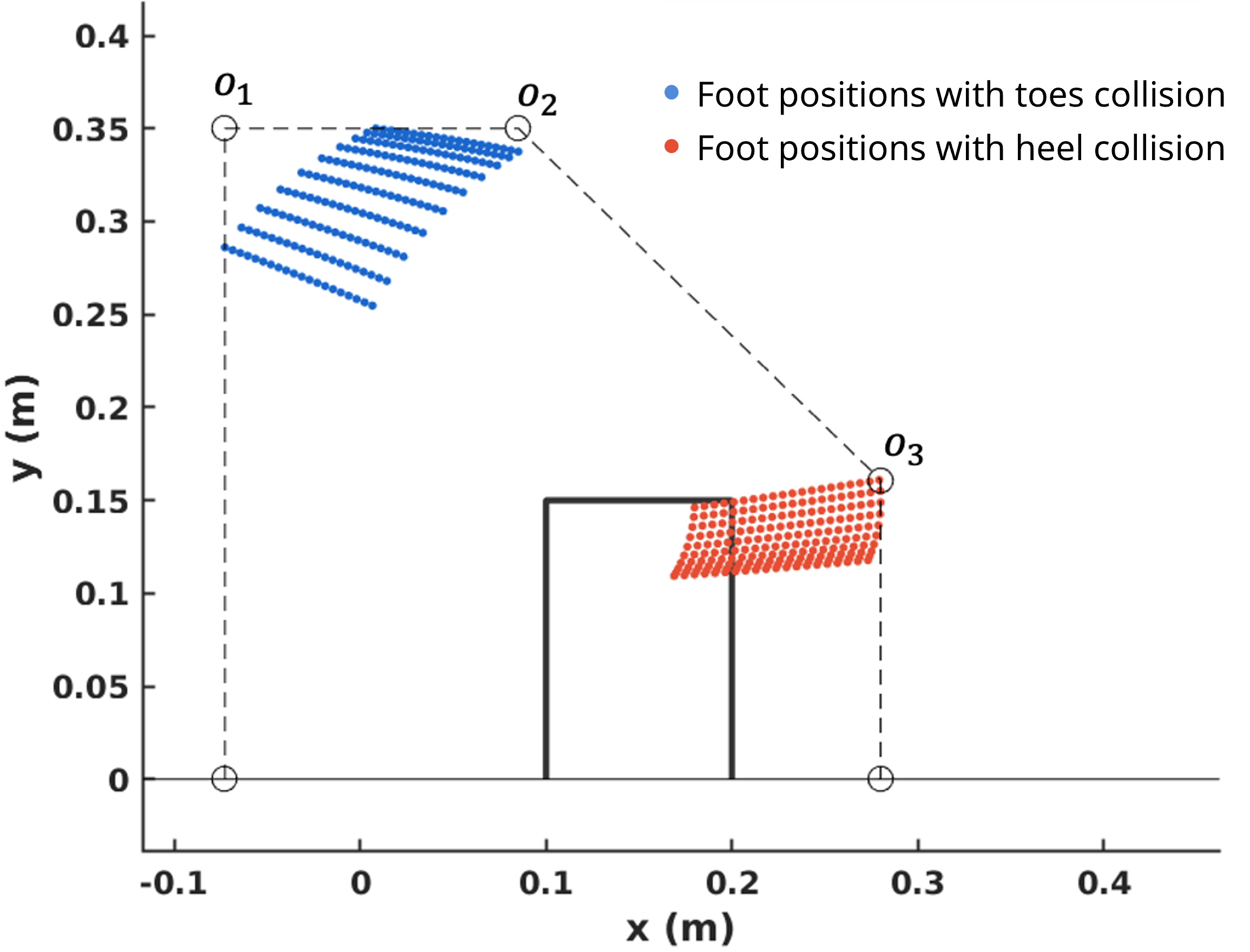}
        \caption{Obstacle's linear constraints identified from the ensemble of foot points for which either the toes (blue dots) or the heel collides with the obstacle surface.}
    \label{fig:enl_fun}
\end{figure}

With these considerations, the proposed algorithm for identifying the obstacle's linear constraints operates as shown in Fig. \ref{fig:enl_fun}: for each point along the CoM trajectory, we identify the possible foot positions $\mathbf{p}_{\text{foot}}=(x_{foot},y_{foot})$ that would result in a contact between the obstacle's surface and either the foot's toes (blue dots), defined as $O^{\text{toes}}$, or the foot's heel (red dots), defined as $O^{\text{heel}}$, as follows:
\begin{equation}
\label{eq:swg_obs_const}
\begin{aligned}
    O^{\text{toes}} = \left\{ \mathbf{p}_{\text{foot}} \in \mathbb{R}^2 \mid \exists i \in \{0, \dots, M-1\} : \mathbf{p}_{\text{toes}} = \mathbf{p}_{o,i} \right\} \\
    O^{\text{heel}} = \left\{ \mathbf{p}_{\text{foot}} \in \mathbb{R}^2 \mid \exists i \in \{0, \dots, M-1\} : \mathbf{p}_{\text{heel}} = \mathbf{p}_{o,i} \right\}
\end{aligned}
\end{equation}
where $\mathbf{p}_{o,i}$ represents the $i$-th point on the obstacle surface. Based on the gait kinematics, we expect possible collisions of the toes in front of the obstacle, and possible collisions with the heel at the obstacle's rear, when the foot is approaching the ground. From these ensembles of identified foot positions, an enlarged bounding box (dotted line) is constructed around the obstacle and described by the points $\{\mathbf{o}_j=(x_{o_j},y_{o_j})\}_{j=1}^{3}$, where $\mathbf{o}_1$ defines the start of the bounding box; $\mathbf{o}_2$ defines the point at which the toes do not pose any collision risk; $\mathbf{o}_3$ defines the point at which the heel does not pose any collision risk. Given these points, we can now define three additional obstacle-related constraints to the swing foot KMP model:
\begin{equation}
\begin{aligned}
    &\mathbf{g}_{n,5}^T = [1,0] \hspace{6mm} c_{n,5}=x_{o_j} \hspace{2mm} : \hspace{2mm} n_{o,j} \le n < n_{o,j+1}\\ 
    &\mathbf{g}_{n,6}^T = [-1,0] \hspace{3mm} c_{n,6}=x_{o_{j+1}} \hspace{2mm} : \hspace{2mm}n \leq n_{o,j} \\ 
    &\mathbf{g}_{n,7}^T = [0,1] \hspace{6mm} c_{n,7}=\frac{y_{o_{j+1}}-y_{o_j}}{x_{o_{j+1}}-{x}_{o_j}} \hat{\mu}_{n,x} \hspace{2mm} : \hspace{2mm} n \leq n_{o,j} \\ 
    &\forall n \in \{1, 2, \dots, N\}, \forall j \in \{1,2,3\}
\end{aligned}
\end{equation}
with
\begin{equation}
    n_{o_{j}} = \min_{r \in \{1,...N\}} (||\hat{\mu}_{r,x} - x_{o_j}||)
\end{equation}
where $\hat{\mu}_{n,x}$ represents the x-coordinate of the reference foot trajectory at index $n$. Intuitively, the above-defined constraints correspond to forcing the generation of the foot trajectory points outside the region defined by the enlarged bounding box. A similar approach as described above is also used in the stair environment to avoid collisions with the edge of the stair tread by adding a virtual obstacle of height equal to the stair tread height and a length of 1 cm.  

Combining all the procedures described in Sections \ref{sec:hybrid_kmp}--\ref{sec:lc_kmp}, the proposed method for the adaptive generation of the swing foot and support leg trajectories is outlined in Algorithm~\ref{alg:kmp_complete}.

\begin{algorithm}[ht]
    \caption{Multi-terrain adaptive gait generation}\label{alg:kmp_complete}
    \begin{algorithmic} [1]
        \STATE Define parameters $l$, $\lambda$ and $\lambda_c$.
        \STATE Collect demonstrations $\{ \{ {s}_{n,h}, \boldsymbol{\xi}_{n,h} \}_{n=1}^{N} \}_{h=1}^{H}$ in $\{O\}$.
        \STATE Compute the reference database $\mathbf{D}=\{{s}_n, \hat{\boldsymbol{\mu}}_n, \hat{\boldsymbol{\Sigma}}_n\}_{n=1}^N$ using GMM and GMR.
        \FOR{Every gait step}
            \STATE \textit{Input:} Receive terrain type, via-points and obstacle points.
            \STATE Define $P=2$ local frames and compute local reference databases $\mathbf{D}^{(p)}=\{{s}_n^{(p)}, \hat{\boldsymbol{\mu}}_n^{(p)}, \hat{\boldsymbol{\Sigma}}_n^{(p)}\}_{n=1}^N$
            \STATE Project via-points and obstacle to local frames (\ref{eq:local_project}).
            \FOR{Each local frame}
                \STATE Update $\mathbf{D}^{(p)}$ with via-points \eqref{eq:update_via}.
                \IF{Swing Foot case}
                    \STATE Add trajectory boundaries \eqref{eq:swg_traj_const} and obstacle linear constraints \eqref{eq:swg_obs_const}.
                    \STATE Set matrices $\overline{\mathbf{G}}$ and $\overline{\mathbf{C}}$ (\ref{eq:matrices_G_C}).
                \ELSIF{Support Leg case}
                    \STATE Add trajectory boundaries constraints \eqref{eq:sup_traj_const}.
                    \STATE Set matrices $\overline{\mathbf{G}}$ and $\overline{\mathbf{C}}$ \eqref{eq:matrices_G_C}.
                \ENDIF
                \STATE Calculate the optimal Lagrange multipliers $\boldsymbol{\alpha}^*$ \eqref{eq:lagrange_multi}.
                \STATE Predict the local trajectory $\tilde{\boldsymbol{\xi}}^{(p)}(\mathbf{s}^{*})$ \eqref{eq:predict_LC_KMP}.
            \ENDFOR
            \STATE \textit{Output:} Compute the final trajectory $\tilde{\boldsymbol{\xi}}(\mathbf{s}^*)$ in the global frame $\{O\}$ \eqref{eq:local_to_global}.
        \ENDFOR
    \end{algorithmic}
\end{algorithm}

\section{Experiments \& Results}
\subsection{Environment Understanding experiments}
This section presents the experimental evaluation of the proposed \textit{Environment Understanding} module, designed to extract relevant geometric properties from RGB-D data. The experiments were conducted in representative environments, including staircases, ramps, and obstacles. In total, 191 independent measurements were collected to compute mean values and associated errors. Quantitative results are summarized in Table~\ref{tab:vision_results}. Overall, the reported results demonstrate the reliability of the proposed \textit{Environment Understanding} module. The method achieves accurate geometric recovery in realistic conditions.

\begin{table}[t] 
\centering 
\caption{Reconstruction error of the \textit{Environment Understanding} module in different environments.}
\renewcommand{\arraystretch}{1.8} 
\setlength{\tabcolsep}{9pt} 
\begin{tabular}{l c c c c c c c} 
\toprule
\textbf{Environment} & & & & & & \\ 
\midrule
\multirow{2}{*}{Stairs} 
& \multicolumn{2}{c}{\textit{Height [cm]}} & \multicolumn{2}{c}{\textit{Length [cm]}} & \multicolumn{2}{c}{\textit{Distance [cm]}} \\ 
\cline{2-7} 
& \multicolumn{2}{c}{$0.20 \pm 0.14$} & \multicolumn{2}{c}{$0.29 \pm 0.19$} & \multicolumn{2}{c}{$0.25 \pm 0.11$} \\ 
\midrule
\multirow{2}{*}{Slopes} 
& \multicolumn{6}{c}{\textit{Inclination [$^\circ$]}} \\
\cline{2-7} 
& \multicolumn{6}{c}{$1.12 \pm 0.37$} \\ 
\midrule
\multirow{2}{*}{Obstacles} 
& \multicolumn{2}{c}{\textit{Height [cm]}} & \multicolumn{2}{c}{\textit{Length [cm]}} & \multicolumn{2}{c}{\textit{Distance [cm]}} \\
\cline{2-7} 
& \multicolumn{2}{c}{$0.42 \pm 0.27$} & \multicolumn{2}{c}{$0.38 \pm 0.41$} & \multicolumn{2}{c}{$0.19 \pm 0.13$} \\
\bottomrule
\end{tabular} 
\label{tab:vision_results}
\end{table} 

\begin{table}[]
\centering
\caption{List of parameters for the swing foot and support leg KMP models.}
\renewcommand{\arraystretch}{1.8} 
\setlength{\tabcolsep}{9pt} 
\begin{tabular}{llllllllll}
\toprule
\textbf{KMP model}&  &  &  & \multicolumn{3}{c}{\textbf{Parameters}}  &  & \\ \midrule
\multirow{2}{*}{{Swing foot}} & & \textit{C} &  & \textit{l} &  & $\lambda$ &  & $\lambda_c$ & \\ \cmidrule(l){3-10}
& & 10 &  & 3 &  & 20 &  & 1 & \\ \midrule
\multirow{2}{*}{Support leg} &  & \textit{C} &  & \textit{l} &  & $\lambda$ &  & $\lambda_c$ & \\ \cmidrule(l){3-9} 
&  & 5 &  & 1 &  & 30 &  & 2 & \\ 
\bottomrule
\end{tabular}
\label{tab:kmp_paramters}
\end{table}

\subsection{Demonstration Dataset \& Parameter Settings}
\begin{figure}[t]
    \centering
    \includegraphics[width=0.5\textwidth]{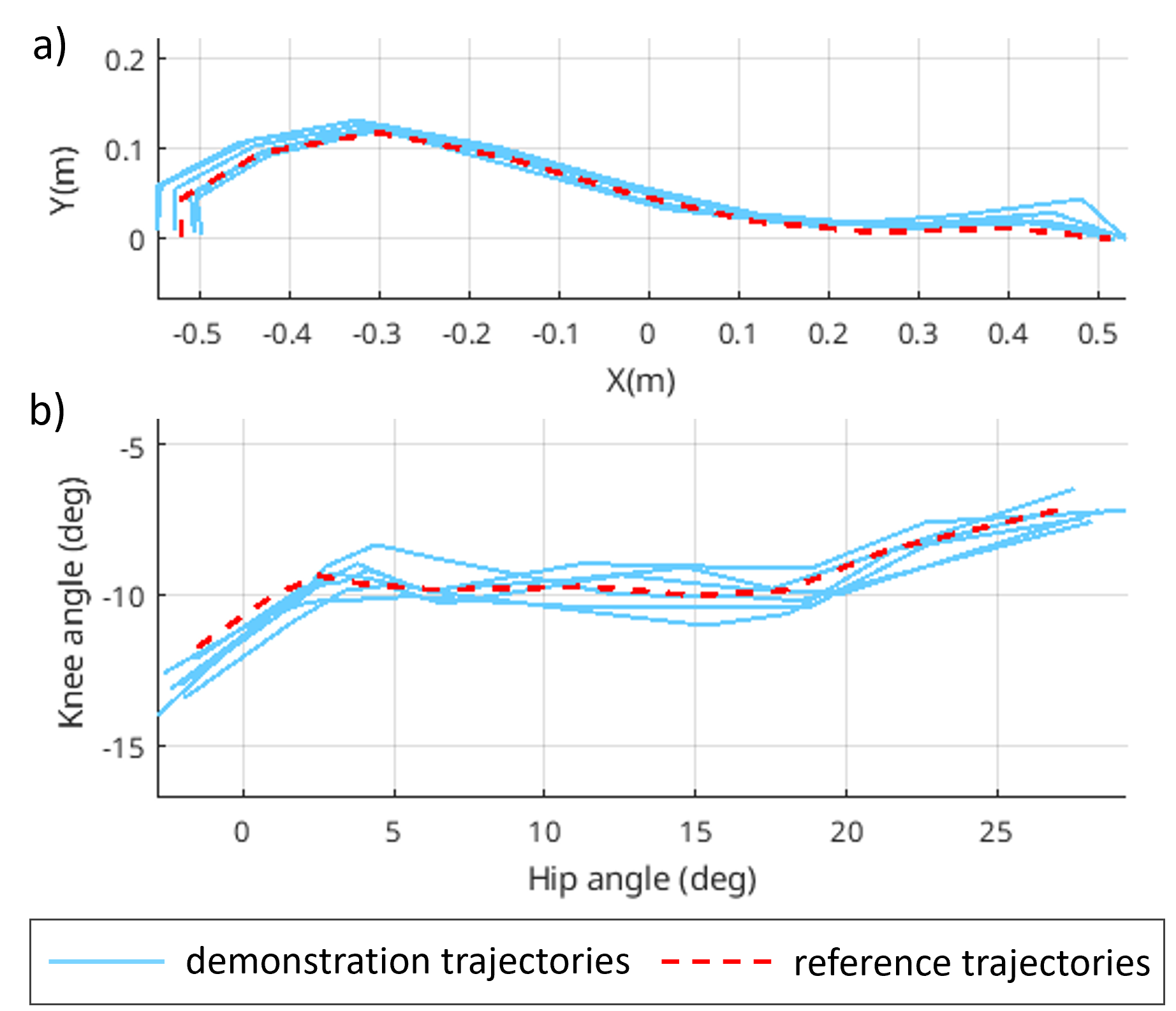}
    \caption{Human demonstrations (blue) used to train the swing foot KMP model (a) and the support leg KMP model (b), together with the reference trajectories (dotted red) generated with the GMR.}
    \label{fig:demos}
\end{figure}

As described in the previous section, the proposed method relies on a set of demonstration trajectories to train the KMP models. In this work, the demonstrations were taken from a publicly available dataset presented in \cite{van_der_zee_biomechanics_2022}, consisting of 33 trials covering various walking conditions from 10 different subjects. In particular, we took the second trial of the first subject (subject 1) as it features walking under preferred conditions—-self-selected step length and frequency--at a walking speed of $0.70\ m.s^{-1}$. These conditions align with the objective of generating gait trajectories for the LLE, which should be as natural as possible and are typically slower than the average human walking speed of $1.31\ m.s^{-1}$ \cite{murtagh2021outdoor}. The subject was instead chosen due to its anthropometric measurements being the closest to the group's mean (25 years old, $81.8\ kg$ weight, $1.73\ m$ height, a leg length of $0.89\ m$, a thigh length of $0.42\ m$, and a shin length of $0.38\ m$). From the selected trial, $H=6$ steps were extracted to form the demonstration dataset, shown in Fig. \ref{fig:demos} (blue lines), with each demonstration comprising $N=10$ sampled points. The corresponding reference trajectories generated by the GMM/GMR models are also shown (dotted red lines) with $C=10$ Gaussian components for the swing foot KMP and $C=5$ for the support leg KMP. The full list of parameters for training the KMP models and generating new trajectories is provided in Table~\ref{tab:kmp_paramters} for both the swing foot and support leg. 

\begin{table*}[t]
\centering
\caption{Pearson's Linear Correlation with physiological data of gait kinematics}
\renewcommand{\arraystretch}{1.3} 
\setlength{\tabcolsep}{20pt} 
    \begin{tabular}{l|cccc}
        \toprule
       \textbf{Simulation} & \textbf{Support Hip} & \textbf{Support Knee} & \textbf{Swing Hip} & \textbf{Swing Knee} \\ \midrule
        Flat-ground - 0.7 m step length & 0.99 & 0.46  & 0.94 & 0.92 \\ \hline
        Flat-ground - 0.92 m step length & 0.99 & 0.43  & 0.98 & 0.93 \\ \hline
        Flat-ground - 1.18 m step length & 0.99 & 0.30  & 1.00 & 0.76 \\ \hline
        Flat-ground - Subject 5 & 0.97 & 0.45  & 0.94 & 0.91 \\ \hline
        Flat-ground - Subject 6 & 0.98 & 0.30  & 0.93 & 0.93 \\ \hline
        Stair ascent - 15.2 cm stair height     & 0.99 & 0.90  & 0.84 & 0.74 \\ \hline
        Stair descent - 15.2 cm stair height & 0.94 & 0.41  & 0.93 & 0.83 \\ \hline
        Slope ascent - 9.2° inclination  & 0.97 & 0.92  & 0.92 & 0.73 \\ \hline
        Slope descent - 9.2° inclination & 0.92 & 0.56  & 0.77 & 0.47 \\ \hline
        \textbf{Average} & $\mathbf{0.97\pm0.02}$ & $\mathbf{0.53\pm0.22}$ & $\mathbf{0.92\pm0.07}$ & $\mathbf{0.80\pm0.14}$ \\ \bottomrule
    \end{tabular}
    \label{tab:corr_walk}
\end{table*}

\begin{figure}[t]
    \centering
    \includegraphics[width=\columnwidth]{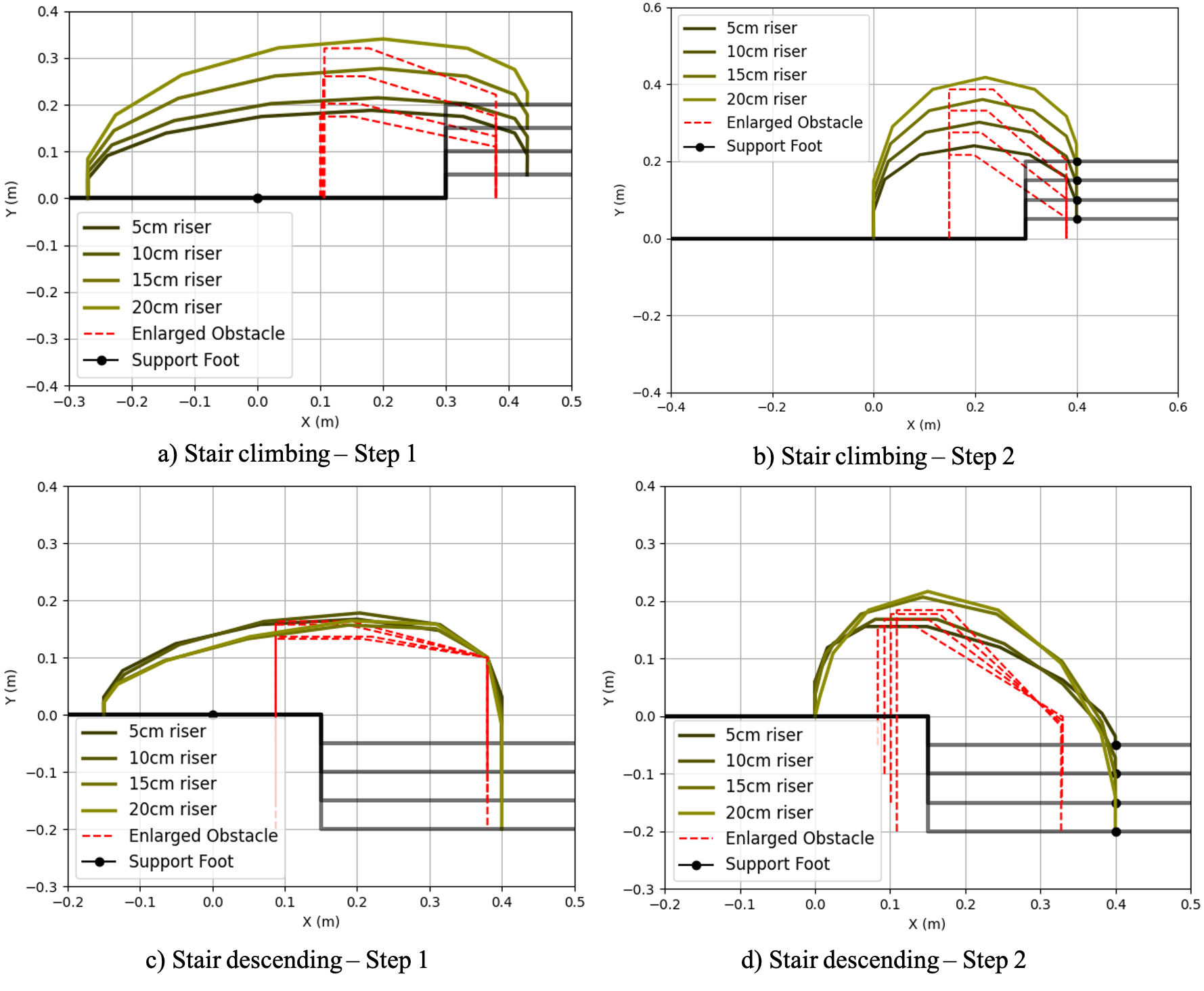}
    \caption{Swing foot trajectory generation for stair ascending (first row) and descending (second row) at different step heights.}
    \label{fig:stairs_multi}
\end{figure}

\begin{figure}[t]
    \centering
    \includegraphics[width=\columnwidth]{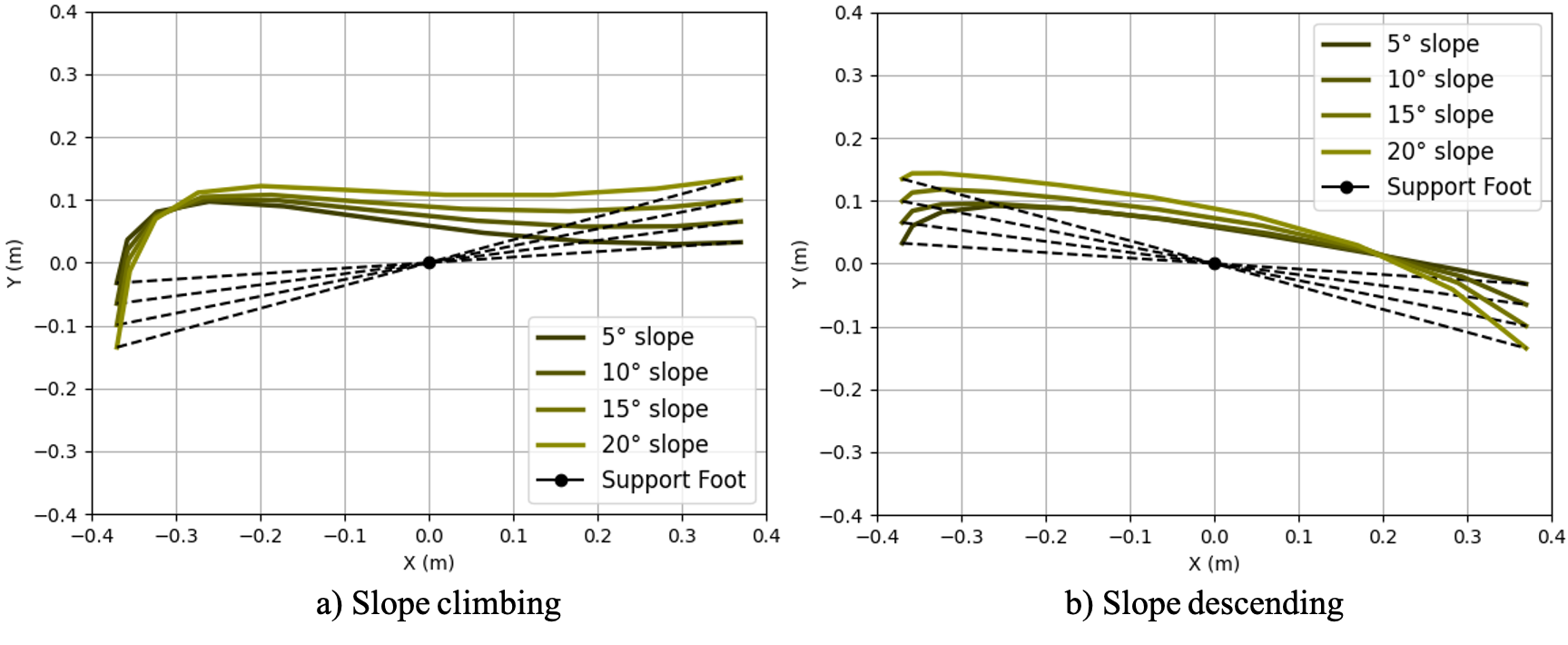}
    \caption{Swing foot trajectories for different slope inclinations.}
    \label{fig:slope_multi}
\end{figure} 

\begin{figure}[t]
    \centering
        \includegraphics[width=\columnwidth]{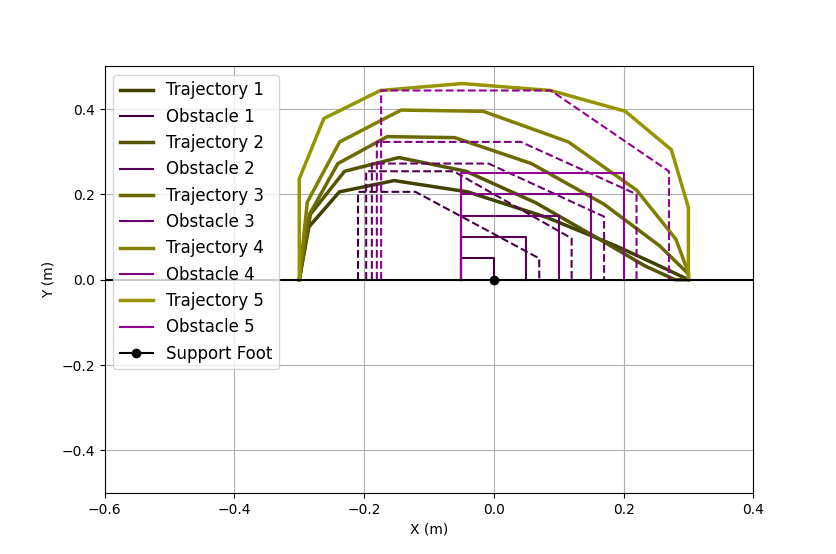}
    
    \caption{Swing foot trajectories for different obstacle sizes (support foot positioned to the side of the obstacle). Scenario 1 comprises an obstacle of 5 x 5 cm, Scenario 2 comprises an obstacle of 10 x 10 cm, up to Scenario 5, which comprises an obstacle of 25 x 25 cm.}
    \label{fig:obs_multi}
\end{figure} 

\subsection{Adaptive Gait Generation Simulations}
For all the considered environments (flat ground, slopes, stairs, and obstacles crossing), gait planning simulations have been conducted in order to verify the validity of the approach before deploying it to a real device. For each simulated scenario, the proposed solution was executed 10 times, and the average trajectory is shown in the figures. To provide a quantitative evaluation of the proposed method, Pearson's Linear Correlation Coefficient \cite{mukaka2012guide} was computed as a measure of trajectory shape similarity between the generated angular trajectories and physiological gait trajectories in similar conditions. The correlation results are displayed in Table \ref{tab:corr_walk}.
\subsubsection{Flat Ground – Variable Step Length}
For the adaptive gait generation on flat ground walking, we compared the generated KMP trajectories with walking data from the same subject used for model training, but with different step lengths of $0.7\ m$, $0.92\ m$, and $1.18\ m$, also available in the public dataset \cite{van_der_zee_biomechanics_2022}.

\subsubsection{Flat Ground – Variable user's parameters}
These simulations were performed to validate the method's capability to generalize to the data of different subjects with different anthropometric characteristics, which were not seen during training. To evaluate this, walking data from two additional subjects in the public dataset \cite{van_der_zee_biomechanics_2022} were selected and compared to the KMP-generated trajectories. In particular, the following subjects were considered:
\begin{itemize}
    \item \textit{Subject 5:} 21 years old, weight $56.7\ kg$, height $1.63\ m$, leg length $0.86\ m$, thigh length $0.40\ m$, shin length $0.36\ m$, and step length $0.96\ m$.
    \item \textit{Subject 6:} 24 years old, weight $72.6\ kg$, height $1.83\ m$, leg length $0.94\ m$, thigh length $0.43\ m$, shin length $0.44\ m$, and step length $0.92\ m$
\end{itemize}

\subsubsection{Stairs}
\label{sec:stairs}
For the adaptive gait generation on stairs, different step heights were considered as shown in Fig \ref{fig:stairs_multi}, for both stair ascent (top) and stair descent (bottom): (i) $5\ cm$, (ii) $10\ cm$, (iii) $15\ cm$, (iv) $20\ cm$. Similarity comparison was performed with respect to physiological gait data from an open-source gait dataset \cite{CAMARGO2021110320} in the same staircase condition (i.e., $15.2\ cm$ step height).

\subsubsection{Slopes}
For the adaptive gait generation on slopes, different inclinations were considered as shown in Fig. \ref{fig:slope_multi}, for both slope ascent (left) and slope descent (right): (i) 5°, (ii) 10°, (iii) 15°, and (iv) 20°. 
Similarly to the stairs scenarios, the generated joint angular trajectories were compared with open-source data \cite{CAMARGO2021110320} in the same environmental condition (i.e., 9.2° slope inclination). 

\subsubsection{Obstacle Crossing}
For the adaptive gait generation with obstacles, we considered simulations of rectangular obstacles with increasing size, from 5 x 5 cm up to 25 x 25 cm, as shown in Fig. \ref{fig:obs_multi}. As shown in the figure, in each trial the step length was kept constant and equal to $0.6\ m$, and only the obstacle size was varied.

\subsection{Experiments with real exoskeleton}
Finally, a full evaluation was also performed on a commercial LLE device. In particular, the exoskeleton platform employed was the UANGO exoskeleton by the company U\&O (Italy), which comprises four active joints (hip and knee of both legs) and two passive joints (ankle of both legs). Each joint is equipped with a linear encoder for measuring the joint position in real time. The adaptive gait planning framework\footnote{Code available at: https://github.com/exoskeleton-iaslab/environment-adaptive-gait-planning.git} was implemented in the Robot Operating System (ROS) \cite{quigley2009ros} running on a mini-PC (Beelink  U59, Intel Celeron 4-Core @ 2.9~GHz, 8~GB~RAM) connected to the exoskeleton low-level controller (i.e., PID motor control) with a custom socket-based communication. The exoskeleton mechanical design was customized to accommodate a Realsense D435 depth camera (Intel Realsense, US) fixed at the exoskeleton pelvis. The length of the links and the size of the pelvis frame were manually adjusted by an expert operator to fit the body characteristics of the user, reported in Table \ref{tab:subj_meas}.

\begin{figure}[t]
    \centering
    \includegraphics[width=\columnwidth]{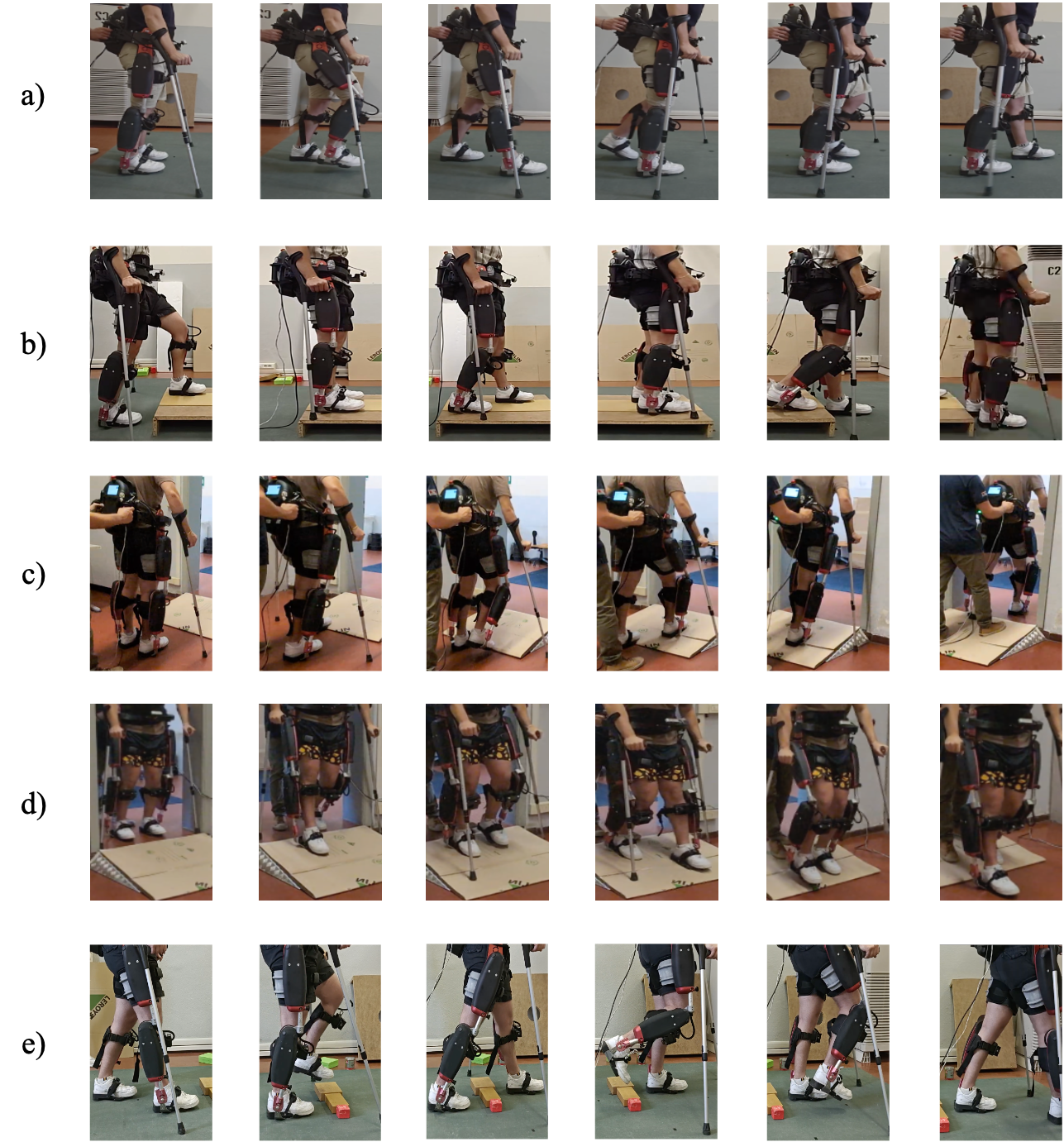}
    \caption{Results of the real experiments in different scenarios (time progression from left to right). a) Flat ground. b) Curb ascending and descending. c) Slope ascending. d) Slope descending. e) Obstacle crossing. }
    \label{fig:real_exps}
\end{figure}

\begin{figure*}[t]
    \centering  \includegraphics[width=\textwidth]{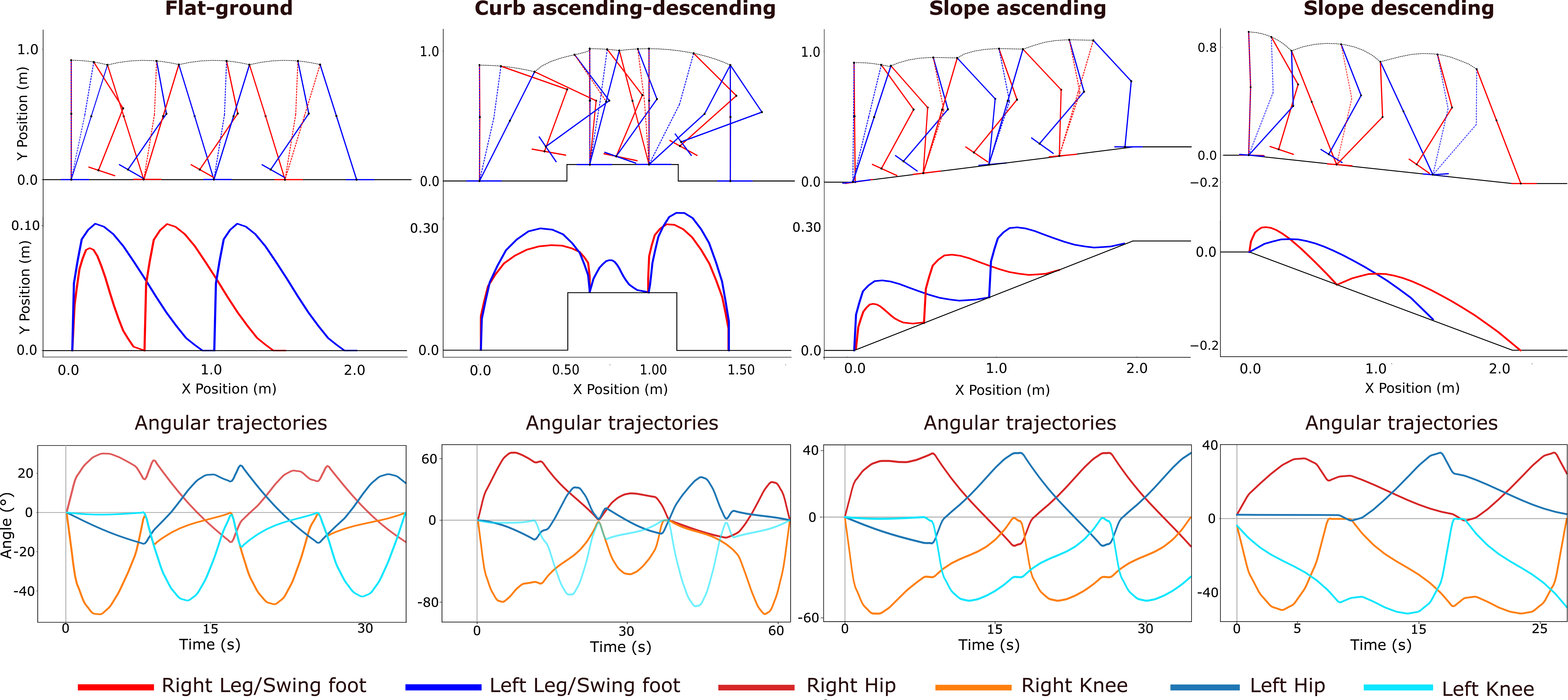}
        \caption{Results of the adaptive gait generation experiments on the real exoskeleton across different terrains (flat-ground, curb ascending-descending, slope ascending and descending). For each terrain, the stick figure and swing foot trajectories are shown (top), together with the generated angular trajectories for hip and knee joints of both legs (bottom).}
    \label{fig:exp_terrains}
\end{figure*}

\begin{figure*}[t]
    \centering  \includegraphics[width=\textwidth]{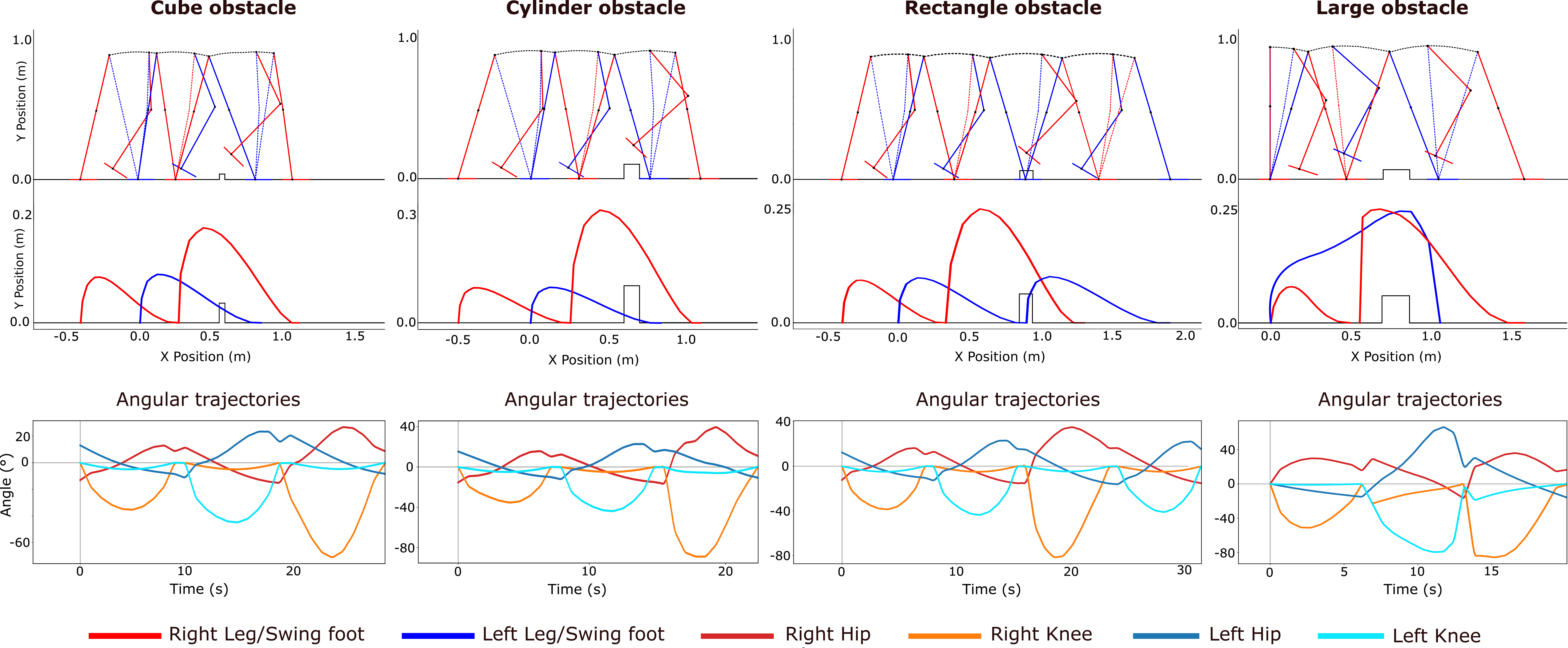}
        \caption{Results of the adaptive gait generation experiments on the real exoskeleton when crossing different types of obstacles (cube, cylinder, rectangle, large obstacle). For each obstacle, the stick figure and swing foot trajectories are shown (top), together with the generated angular trajectories for hip and knee joints of both legs (bottom).}
    \label{fig:exp_obs}
\end{figure*}

The subject was requested to perform multiple steps in the following scenarios:
\begin{enumerate}
    \renewcommand{\labelenumi}{\roman{enumi}.}
    \item Flat ground walking:
    \begin{itemize}
        \item Step Length: $30\ cm$, Step Height: $10\ cm$ 
        \item Step Length: $50\ cm$, Step Height: $10\ cm$ 
        \item Step Length: $50\ cm$, Step Height: $20\ cm$ 
        \item Step Length: $50\ cm$, Step Height: $30\ cm$ 
        \item Step Length: $70\ cm$, Step Height: $10\ cm$ 
    \end{itemize}
    \item Slope ascending (6° inclination) 
    \item Slope descending (6° inclination)  
    \item Curb ascending and descending  ($13\ cm$ height)
    \item Obstacle crossing:
    \begin{itemize}
        \item Cube obstacle: length $5\ \text{cm}$, width $5\ \text{cm}$, height $5\ \text{cm}$
        \item Cylinder obstacle: height $10\ \text{cm}$, diameter $10\ \text{cm}$
        \item Rectangular obstacle: length $16\ \text{cm}$, width $16\ \text{cm}$, height $5\ \text{cm}$
        \item Large obstacle: length $15\ \text{cm}$, width $40\ \text{cm}$, height $5\ \text{cm}$
    \end{itemize}
\end{enumerate} 
All trials have been repeated at least twice to further verify the robustness of the proposed solution. Snapshots of some performed trials are displayed in Fig. \ref{fig:real_exps}, while the generated trajectories are shown in Fig. \ref{fig:exp_terrains} for the different terrains, and in Fig. \ref{fig:exp_obs} for each obstacle crossing scenarios.
Every trial has been completed without failures or collisions, demonstrating the validity of the proposed approach. Additionally, we compared the generated angular trajectories for each joint with the effective angular trajectories executed by the exoskeleton and measured by the encoders. Overall, we found a mean Root Mean Square Error (RMSE) in the execution of the trajectories of $3.74^{\circ}$ and an average computation time for generating the trajectories of $0.57\ s$ (see Table \ref{tab:exp_rmse}).

\begin{table}[]
    \centering
    \caption{Anthropometric measurements of the subject in the real exoskeleton experiments}
    \renewcommand{\arraystretch}{1.3} 
\setlength{\tabcolsep}{25pt} 
    \begin{tabular}{l|c|c}
    \toprule
         Thigh length & $L_1$ & $0.46\ m$ \\
         Shin length & $L_2$ & $0.4\ m$ \\
         Foot height & $L_3$ & $0.09\ m$ \\ 
         Heel-to-foot length & $L_4$ & $0.08\ m$ \\
         Foot-to-toes length & $L_5$ & $0.2\ m$      
    \end{tabular}
    \label{tab:subj_meas}
\end{table}

\begin{table}[h]
\centering
\caption{Algorithm computation time and RMSE between planned and measured joint trajectories, for each environment and specific condition, averaged across multiple trials}
    \begin{tabular}{|l|p{2.3cm}|c|c|}
        \hline
       \textbf{Environment} & \textbf{Condition} & \textbf{RMSE($^{\circ}$)} & \textbf{Time(s)}  \\ \hline
        Flat Ground  & Step Length: 0.3 m \newline Step Height: 0.1 m  & $3.69 \pm 0.058 $  & $0.39 \pm 0.005$ \\ \hline
        Flat Ground  & Step Length: 0.5 m \newline Step Height: 0.1 m  & $3.63 \pm 0.063$ & $0.39 \pm 0.005$ \\ \hline
        Flat Ground  & Step Length: 0.5 m \newline Step Height: 0.2 m  & $3.63 \pm 0.034$ & $0.39 \pm 0.005$ \\ \hline
        Flat Ground  & Step Length: 0.5 m \newline Step Height: 0.3 m  & $3.59 \pm 0.053$ & $0.39 \pm 0.005$  \\ \hline
        Flat Ground  & Step Length: 0.7 m \newline Step Height: 0.1 m  & $3.53 \pm 0.037$ & $0.39 \pm 0.005$ \\ \hline
        Curb  & Ascending and \newline Descending  & $3.30 \pm 0.43$  & $0.55 \pm 0.173$\\ \hline
        Slope  & Ascending   & $3.43 \pm 0.30$  & $0.4 \pm 0.015$\\ \hline
        Slope  & Descending   & $3.18 \pm 2.02$  & $0.39 \pm 0.003$\\ \hline
        Obstacle  & Cube   & $2.88 \pm 0.18$ & $0.83 \pm 0.078$ \\ \hline
        Obstacle  & Cylinder   & $3.30 \pm 0.21$ & $0.9 \pm 0.035$ \\ \hline
        Obstacle  & Rectangle   & $3.81 \pm 0.39$  & $0.92 \pm 0.013$\\ \hline
        Obstacle  & Large Obstacle   & $4.78 \pm 0.54$  & $0.87 \pm 0.021$\\ \hline
        
    \end{tabular}
    \label{tab:exp_rmse}
\end{table}

\section{Discussion and Conclusions}
In this paper, a vision-based multi-terrain adaptive gait generation system for exoskeletons is proposed. The system features an automatic terrain recognition and characterization, and a novel adaptive gait generator capable of producing at run-time safe and physiological plausible trajectories adapted to the specific environment.

To achieve an appropriate terrain-adaptive exoskeleton control, an accurate and timely recognition of the environment is crucial. To tackle this challenge, we developed a vision-based approach based on depth images with high accuracy, as demonstrated by the experimental results (Table \ref{tab:vision_results}). In particular, the MLESAC-based multi-plane segmentation strategy proved effective in managing noisy and cluttered point clouds, enabling reliable detection of stairs and slopes with a reconstruction error which is lower than similar methods at the state of the art \cite{zhao2019adaptive, guo2024terrain, zhang2025vision}. Additionally, our \textit{Environment Understanding} module is capable of dealing with unstructured elements that cannot be traced back to specific terrain types (i.e., flat-ground, stairs, slopes) by treating them as obstacles to be avoided. Compared to our previous method \cite{trombin2024environment}, the reconstruction accuracy of the obstacles' shape and position is further improved, and outperforms other RANSAC-based methods in the literature~\cite{liu_vision-assisted_2021, hua_vision_2022}. These results validate the ability of the perception module to provide consistent and contextual information for safe and adaptive gait planning in real-world home environments. 

Regarding the AGG module, the extensive validation experiments in simulations demonstrate the effectiveness and the generalization capabilities of the proposed approach to deal with a high number of different terrain configurations. Although the KMP models were trained exclusively on level-ground walking demonstrations, the method still produced feasible gait trajectories across all indoor environments. Even if an explicit quantitative comparison with other methods in the literature is not possible due to the lack of standard benchmarks and open-source code and data, herein we propose to measure the goodness of the generated trajectories through a similarity comparison with physiological gait data (Table \ref{tab:corr_walk}). The results revealed a strong consistency between physiological and generated trajectories in all the terrains, with an average correlation $>0.90$ for the hip joints, and of $0.80$ for the knee joint of the swing leg. This analysis is useful not only to provide a baseline for the research community to compare with, but also to identify direction of improvements for future work. In particular, the results show that the knee of the support leg represents the most complicated joint to reconstruct, with an average correlation of $0.53$ and a high variability across the terrains. This can be explained by the fact that our model is focused on the kinematic aspects of the human locomotion. However, during the stance phase the human knee joint functions principally as a shock absorber for weight acceptance, working as a linear torsional spring with suitable stiffness \cite{chen2019knee}. The stiffness of the joint is regulated based on the individual's body size, terrain, and gait conditions \cite{shamaei2013estimation}, a behavior that is not explicitly encoded in our model. 

The proposed AGG method posses also generalization capabilities to different users. Even if the KMP models are trained on the data of a specific subject, our results demonstrate that the method can generate walking patterns with a high correlation (i.e., $>0.90$) to the natural walking motion of unseen subjects with different anthropometric measurements and walking characteristics (see Table \ref{tab:corr_walk}). The same KMP models were also directly used with our user for the control of the real exoskeleton without the need of model's re-calibration. This highlights from the one hand the user-independent capability of our AGG system, but also the independence to the specific robotic device, as no data from the real exoskeleton has been used during the system training. This aspect allows to apply our system to different exoskeleton platforms and to new users directly without any data collection or parameter tuning, removing a potential barrier for the effective use in a real application with end-users.

The experimental tests with a real exoskeleton platform represents a fundamental step for demonstrating the effectiveness and robustness of the gait generator to the non-ideality and variability of complex real-world scenarios or inaccuracies of the vision system, which is sometimes omitted in the literature \cite{trombin_environment-adaptive_2025, yang2024adaptive,huang2020adaptive, ma_gait_2018}. In all the experimental tests, both the task space constraints (e.g., obstacles, riser plane of stairs) and angular constraints (e.g., joints angular limits) were consistently satisfied, resulting in a safe and smooth walking across a large variety of indoor scenarios. Indeed, differently from existing studies that focus on a specific terrain \cite{trombin2024environment, huang_adaptive_2020, chen_learning_2019, ma_gait_2018, zou_adaptive_2019, yang2024adaptive, zhao2019adaptive} or a limited set of terrains \cite{mohamad2023online, zou_terrain-adaptive_2023, guo2024terrain}, our work is the first to consider and successfully test a vision-based AGG system with a real exoskeleton in all the most common indoor settings. In particular, while several works exists in the literature for the stair ascending task \cite{mohamad_online_2023,kittisares2023ergonomic,zhao2019adaptive,raineri_adaptive_2024,zhao2019adaptive} also with imitative methods \cite{ma_gait_2018,yang2024adaptive,zhang_study_2024,chen_learning_2019}, to the best of our knowledge this is the first work that considers the stair descending scenario. Additionally, the same multi-terrain framework seamlessly incorporates the collision avoidance of unstructured obstacles in the environment, a challenge often neglected in the field.

Finally, for the effective deployment of the system on a wearable robot, it is important to consider the computational load given the limited computing power of on-board processing units. Specifically, for processing the point cloud and generating the gait trajectories, our system takes on average about $400\ ms$ in all the terrains (Table \ref{tab:exp_rmse}). A higher computation time is required for the obstacle crossing scenarios. This increase is principally due to the time taken by the clustering algorithm and by the identification of a higher number of collision points for the creation of the bounding box in the collision avoidance constraints. Nevertheless, also in the most complicated scenarios the generation time is consistently lower than $1\ s$, which is sufficient for the usual step cadence of commercial lower limb exoskeletons (i.e., $<1\ Hz$) \cite{louie2015gait}.

In conclusion, this work presents a novel method for the generation of adaptive gait in walking exoskeletons suitable for indoor applications. Future work will focus, from the one hand, on extending the perception capabilities of the exoskeleton including machine-learning based classification methods for terrain recognition \cite{chen2022unsupervised} and visual-inertial odometry capabilities \cite{wang_exosense_2024}. On the other hand, we will extend the adaptive gait generation to multi-step planning in order to anticipate the exoskeleton behavior several steps ahead and improve the safety of the system in cluttered environments. Nevertheless, both the simulation results and the experiments on a real commercial device supports the effectiveness of our method to pave the way for a new generation of robotic exoskeletons for the daily assistance of people with disabilities.

\section*{Acknowledgment}
We acknowledge funding from PE00000013 “PNRR MUR - M4C2 - "Future Artificial Intelligence Research- FAIR” - SPOKE 5 - CUP B53C22003980006, in the context of the National Recovery and Resilience Plan, Mission 4, Component 2, Investment 1.3 financed by the European Union, NextGenerationEU. The work is also partially supported by Next Generation EU, in the context of the NRRP PE8 – Project Age-It: “Ageing Well in an Ageing Society” [DM 1557 11.10.2022].


%





\ifCLASSOPTIONcaptionsoff
  \newpage
\fi





\bibliographystyle{IEEEtran}
\bibliography{IEEEabrv,Bibliography}


\vfill


\end{document}